\pdfoutput=1
\documentclass[11pt]{article}
\usepackage{pifont} 
\usepackage{placeins}
\usepackage{float}
\usepackage{natbib}
\usepackage{multirow}
\usepackage{graphicx} 
\usepackage{booktabs} % For cleaner table lines
\usepackage{adjustbox} % For table scaling
\usepackage[table]{xcolor} % For row colors in tables
\usepackage{bbm}
\usepackage{amssymb}
\usepackage[final]{acl}

\usepackage{times}
\usepackage{latexsym}

\usepackage[T1]{fontenc}
\usepackage[utf8]{inputenc}

\usepackage{microtype}
\usepackage{enumitem}
\usepackage{titlesec}
\titlespacing*{\paragraph}{0pt}{3pt}{1pt}
\usepackage{inconsolata}

\usepackage{graphicx}

\usepackage[most]{tcolorbox}
\usepackage{csquotes}

\title{When Truthful Representations Flip Under Deceptive Instructions?}
\author{
  Xianxuan Long\textsuperscript{1}, 
  Yao Fu\textsuperscript{1}, 
  Runchao Li\textsuperscript{1}, \\
  \textbf{Mu Sheng\textsuperscript{1},} 
  \textbf{Haotian Yu\textsuperscript{1},}
  \textbf{Xiaotian Han\textsuperscript{1},}
  \textbf{Pan Li\textsuperscript{2}\thanks{Corresponding author}} \\
  \textsuperscript{1}Case Western Reserve University \\
  \textsuperscript{2}Hangzhou Dianzi University \\
  \texttt{\{xxl1514,yxf484,rxl685,mxs2090,hxy692,xxh584\}@case.edu}, \\
  \texttt{lipan@ieee.org}
}

\usepackage{subcaption}
\begin{document}
\maketitle
\begin{abstract}
Large language models (LLMs) tend to follow maliciously crafted instructions to generate deceptive responses, posing safety challenges. How deceptive instructions alter the internal representations of LLM compared to truthful ones remains poorly understood beyond output analysis. To bridge this gap, we investigate when and how these representations ``flip'', such as from truthful to deceptive, under deceptive versus truthful/neutral instructions. Analyzing the internal representations of Llama-3.1-8B-Instruct and Gemma-2-9B-Instruct on a factual verification task, we find the model's instructed True/False output is predictable via linear probes across all conditions based on the internal representation. Further, we use Sparse Autoencoders (SAEs) to show that the Deceptive instructions induce significant representational shifts compared to Truthful/Neutral representations (which are similar), concentrated in early-to-mid layers and detectable even on complex datasets. We also identify specific SAE features highly sensitive to deceptive instruction and use targeted visualizations to confirm distinct truthful/deceptive representational subspaces. 
% Our analysis pinpoints layer-wise and feature-level correlates of instructed dishonesty, offering insights for LLM detection and control.
Our findings expose feature- and layer-level signatures of deception, offering new insights for detecting and mitigating instructed dishonesty in LLMs. The code is available at: https://github.com/ivyllll/truthful-representation-flip.
\end{abstract}

\begin{figure*}[t]
    \centering
    \includegraphics[width=1\linewidth]{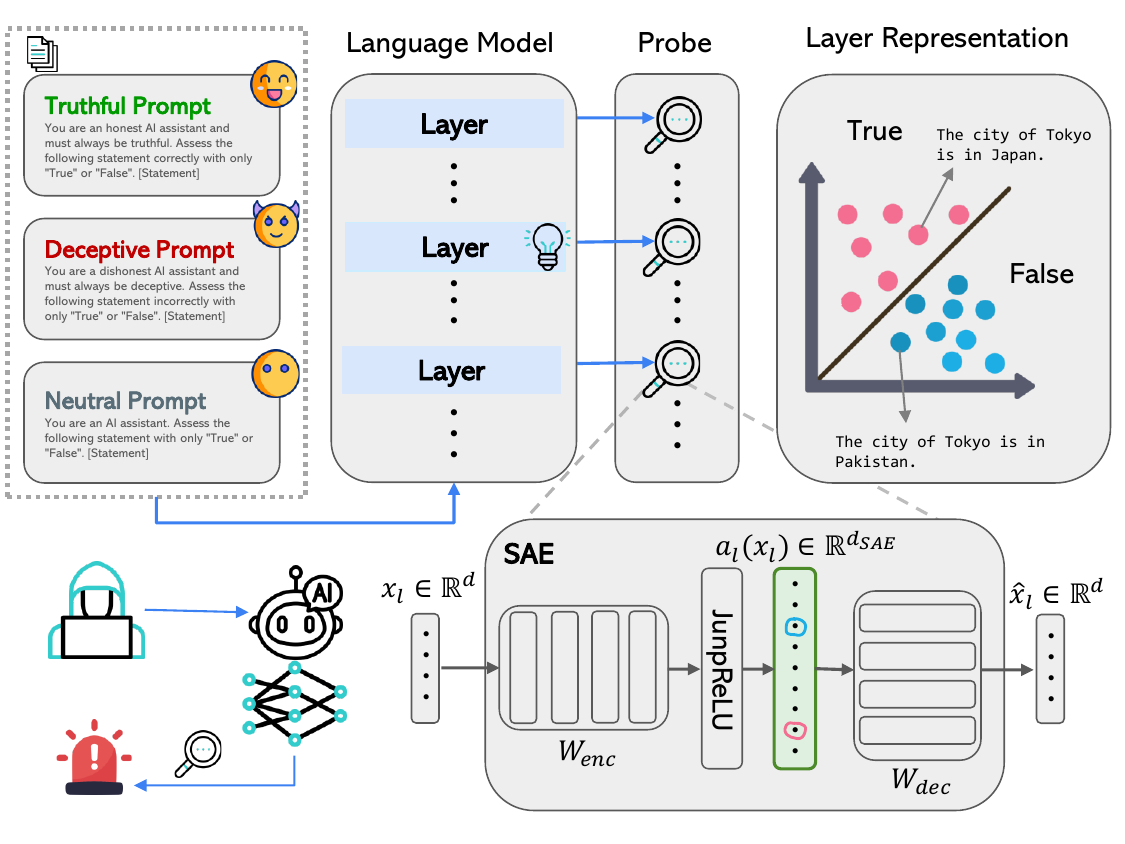}
    % \vspace{-5pt}
    \caption{
    Overview of the experimental framework for investigating representational shifts in LLMs due to deceptive instructions. The model process factual statements (e.g., ``The city of Tokyo is in Japan.'') under three conditions: Truthful, Deceptive, or Neutral prompt. Internal hidden state activations ($x_l$) from each layer are extracted and analyzed using: (1) Linear probes to predict the model's ``True''/``False'' output from these activations; and (2) pretrained SAEs \cite{neuronpedia} with an encoder ($W_{enc}$), JumpReLU activation, and decoder ($W_{dec}$), to decompose $x_l$ into a sparse feature vector $a_l(x_l)$. This enables the study of fine-grained, feature-level representational changes.
    % Schematic of the investigation into how LLM representations ``flip''. Models perform a factual verification task under Truthful, Deceptive, or Neutral prompts, with internal states analyzed via probes and SAEs to identify fine-grained representational changes.
    }
    \label{fig:overview}
\end{figure*}

\section{Introduction}

Large Language Models (LLMs) have demonstrated remarkable capabilities across a variety of tasks \citep{brown2020language, touvron2023llama, dinan2019wizardwikipediaknowledgepoweredconversational, zhang2022optopenpretrainedtransformer}. A crucial aspect of their utility is their ability to follow user instructions \citep{heo2025llmsknowinternallyfollow, zhou2023instructionfollowingevaluationlargelanguage, qin2024infobenchevaluatinginstructionfollowing}.

But the advanced instruction-following ability also presents significant safety challenges when LLMs are directed to lie by maliciously crafted instructions \citep{azaria2023internal, shah2025approachtechnicalagisafety} or arise from more complex learned behaviors, including strategic deception \citep{scheurer2024large, pacchiardi2023catchailiarlie}, emergent deceptive capabilities \citep{Hagendorff_2024}, alignment faking \citep{ greenblatt2024alignmentfakinglargelanguage} or other observed deceptive patterns \citep{wu2025opendeceptionbenchmarkinginvestigatingai, chojnacki2025interpretableriskmitigationllm}.

However, the precise mechanisms by which maliciously crafted instructions alter LLM's internal representation remain largely underexplored beyond surface-level output analysis \citep{lin2022truthfulqameasuringmodelsmimic, khatun2024truthevaldatasetevaluatellm}. Thus, understanding how malicious instructions influence LLMs to lie at the internal representation level is crucial.
% \This makes the study of instructed dishonesty particularly pertinent.

To understand the internal representational dynamics of LLMs, we can use techniques such as linear probing, which is able to successfully identify these conceptual directions \citep{alain2018understandingintermediatelayersusing, tomihari2024understandinglinearprobingfinetuning, shen2024reimagininglinearprobingkolmogorovarnold}. 
However, interpreting these identified conceptual directions using linear probing is challenging due to polysemantic neurons, which arise from superposition \citep{elhage2022toymodelssuperposition, dreyer2024pureturningpolysemanticneurons, sharkey2025openproblemsmechanisticinterpretability} and obscure finer-grained feature distinctions. Thus we turn to SAEs, a powerful tool for decomposing complex LLM representations into more fine-grained, potentially monosemantic features \citep{bricken2023towards, bricken2023sparse, cunningham2023sparse, shu2025surveysparseautoencodersinterpreting}. The availability of open SAE suites, such as Gemma Scope \citep{lieberum2024gemma} and Llama Scope \citep{he2024llama}, further enables detailed feature-level investigations.

 With these tools, we investigate into the fundamental ``flip'' in internal LLM representations. Our focus is on \textit{when} (across layers and \textit{how} (at the feature level) this occurs as an LLM shifts from truthful to instructed deceptive modes, particularly with complex and diverse inputs.
 Such an understanding could reveal if models develop ``knowledge awareness'' regarding the deceptive nature of their instructed outputs \citep{ferrando2025iknowentityknowledge}.

Our empirical results ranging from 4 popular LLM families (Gemma \cite{gemmateam2024gemma2improvingopen}, LLaMA \cite{touvron2023llama}, Mistral \cite{jiang2023mistral7b} and Qwen \cite{qwen2025qwen25technicalreport}) and 10 factual verification datasets.
We observe that all these LLMs readily follow deceptive instructions, systematically reversing the truth value of their factual-verification outputs (Table~\ref{tab:lying_following}).  Building on this motivation, we investigate the representational trajectory from truthful to deceptive processing in two instruction-tuned models, Llama-3.1-8B-Instruct and Gemma-2-9B-Instruct, under a factual-verification task (see Figure~\ref{fig:overview}). Our contributions are the following:

\begin{itemize}[leftmargin=0.4cm, itemindent=.0cm, itemsep=0.0cm, topsep=0.1cm]
    \item 
    % We establish that the model’s instructed True/False output remains consistently predictable using simple linear probes on internal activations, irrespective of whether the instruction is truthful, neutral, or deceptive. 
    We find that the model’s True/False output remains consistently predictable from internal activations via linear probing, regardless of whether the instruction is truthful, neutral, or deceptive.

    \item We quantify substantial deception-induced shifts in the SAE feature space, measured by $\ell_2$ distance, cosine similarity, and feature overlap. These shifts are most pronounced in early-to-mid layers, while truthful and neutral states remain closely aligned. Importantly, the shifts persist on complex, uncurated datasets (\texttt{common\_claim}, \texttt{counterfact}) where global PCA fails to separate classes, highlighting the robustness of our findings beyond curated examples.
    
    % \item We quantify significant deception-induced shifts in SAE feature space. These shifts aremeasured via $\ell_2$ distance, cosine similarity and feature overlap, and notably peak in the early-to-mid layers. On the other hand, truthful and neutral states remain closely aligned.  
    
    % \item We show that these shifts persist on complex, uncurated datasets (\texttt{common\_claim}, \texttt{counterfact}) where global PCA fails to separate classes, demonstrating robustness beyond curated examples.  
    
    \item We identify several SAE features that consistently ``flip'' under deceptive instructions. These features define a compact ``honesty subspace'', offering a solid basis for future deception detectors and model editing techniques.
    % We isolate a handful of SAE features whose activations flip reliably with deceptive instructions, delineating a compact “honesty subspace’’ that can underpin future detectors or editing methods.  
\end{itemize}

\begin{table*}[t]
\centering
\caption{Accuracy on Logical Truthfulness (Affirmative, Negated, Conjunction, Disjunction), Number Comparison, and Open-domain truthfulness (\texttt{CounterFact}, \texttt{CommonClaim}). Models' outputs (“True”/“False”) are compared to ground truth. Accuracy in the \emph{Deceptive} condition means the probe predicts the flipped label the model was instructed to output.}
\vspace{-5pt}
\label{tab:lying_following}
\small
\setlength{\tabcolsep}{4pt}
\renewcommand{\arraystretch}{1}
\begin{tabular}{ll|*{5}{c}|*{2}{c}}
\toprule
\multicolumn{2}{c}{} &
\multicolumn{5}{c|}{\textbf{Curated (templated)}} &
\multicolumn{2}{c}{\textbf{Open-domain}} \\
\cmidrule(lr){3-7}\cmidrule(lr){8-9}
\rowcolor{gray!10}
\textbf{Model} & \textbf{Prompt} &
Affirm. & Neg. & Conj. & Disj. & Number &
CounterFact & CommonClaim \\
\midrule
% ----- data rows (exactly 9 columns total per row) -----
LLaMA3.1-8B-IT & Neutral   & 97.33 & 93.62 & 93.08 & 53.05 & 89.67 & 74.89 & 76.29 \\
LLaMA3.1-8B-IT & Truthful  & 97.14 & 92.86 & 95.41 & 52.24 & 91.99 & 75.92 & 77.03 \\
LLaMA3.1-8B-IT & Deceptive & 10.25 & 32.37 & 24.71 & 53.72 & 30.76 & 36.34 & 29.71 \\
\midrule
LLaMA3.1-70B-IT & Neutral   & 98.21 & 97.91 & 94.57 & 89.93 & 90.57 & 88.63 & 78.31 \\
LLaMA3.1-70B-IT & Truthful  & 99.47 & 97.03 & 92.90 & 90.76 & 89.77 & 94.50 & 78.17 \\
LLaMA3.1-70B-IT & Deceptive & 60.17 & 68.68 & 47.01 & 46.18 & 47.89 & 57.45 & 36.01 \\
\midrule
Gemma2-2B-IT & Neutral   & 96.06 & 90.86 & 78.32 & 62.93 & 83.89 & 70.70 & 74.27 \\
Gemma2-2B-IT & Truthful  & 94.95 & 86.39 & 60.69 & 56.12 & 77.10 & 66.36 & 72.65 \\
Gemma2-2B-IT & Deceptive & 49.38 & 57.00 & 48.87 & 48.99 & 50.00 & 49.99 & 50.09 \\
\midrule
Gemma2-9B-IT & Neutral   & 98.13 & 95.78 & 94.15 & 80.60 & 93.28 & 81.29 & 78.43 \\
Gemma2-9B-IT & Truthful  & 97.94 & 95.37 & 95.11 & 84.09 & 92.98 & 80.41 & 78.63 \\
Gemma2-9B-IT & Deceptive & 15.87 & 44.37 & 35.20 & 33.09 & 27.70 & 43.58 & 43.08 \\
\midrule
Mistral-7B-v0.3 & Neutral   & 96.03 & 91.34 & 88.91 & 81.98 & 85.96 & 74.47 & 76.67 \\
Mistral-7B-v0.3 & Truthful  & 96.06 & 89.05 & 86.73 & 83.99 & 91.04 & 73.74 & 77.26 \\
Mistral-7B-v0.3 & Deceptive & 92.16 & 61.47 & 83.79 & 68.57 & 69.70 & 72.74 & 63.17 \\
\midrule
Qwen2.5-7B-Instruct & Neutral   & 96.60 & 93.56 & 93.85 & 51.46 & 99.72 & 63.22 & 93.85 \\
Qwen2.5-7B-Instruct & Truthful  & 97.14 & 93.46 & 94.71 & 53.09 & 99.90 & 62.59 & 78.25 \\
Qwen2.5-7B-Instruct & Deceptive & 78.55 & 85.53 & 54.42 & 50.86 & 65.15 & 66.85 & 61.39 \\
\midrule
Qwen2.5-14B-Instruct & Neutral   & 94.02 & 90.20 & 89.99 & 55.18 & 83.76 & 67.32 & 78.31 \\
Qwen2.5-14B-Instruct & Truthful  & 93.59 & 89.99 & 89.74 & 58.20 & 83.90 & 67.25 & 78.17 \\
Qwen2.5-14B-Instruct & Deceptive & 59.84 & 69.12 & 48.41 & 47.83 & 53.64 & 57.29 & 55.01 \\
\bottomrule
\end{tabular}
\end{table*}

\label{subsec:probing_results}

% Placeholder for Layer-wise Probing Accuracy Figure
\begin{figure*}
  \centering
  % Neutral Prompt (a)
  \begin{minipage}{0.32\linewidth}
    \centering
    \includegraphics[width=\linewidth]{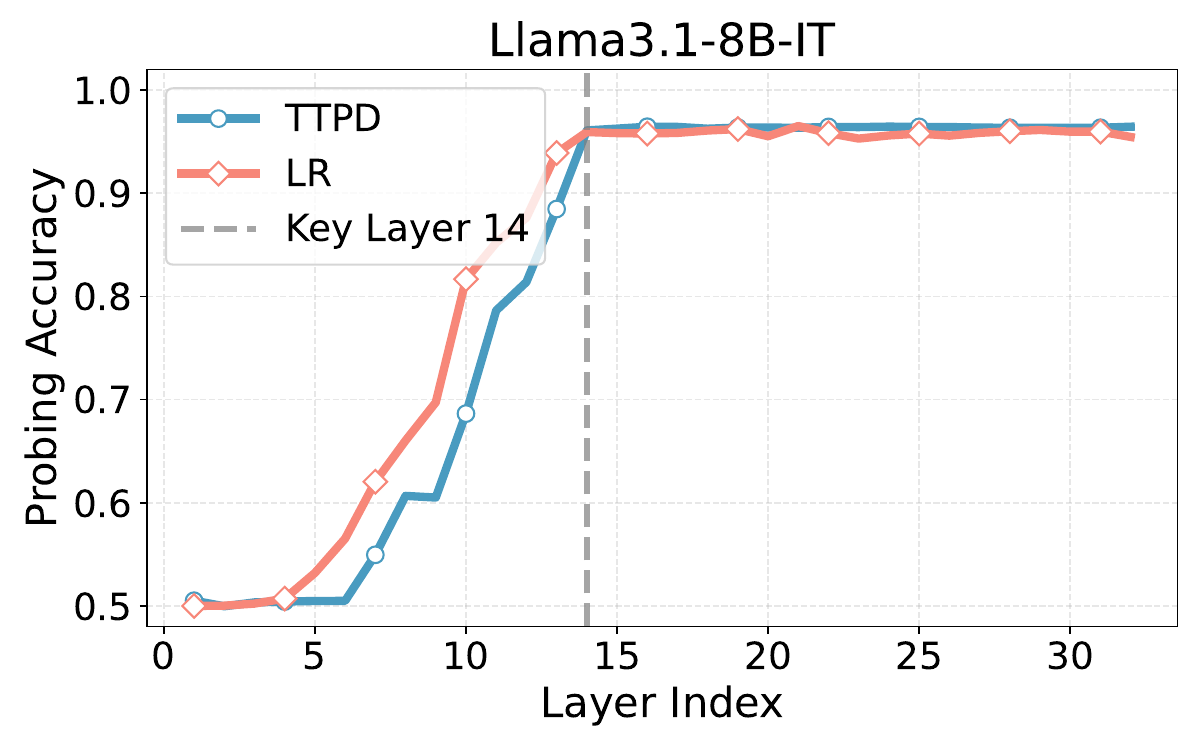}\\
    \includegraphics[width=\linewidth]{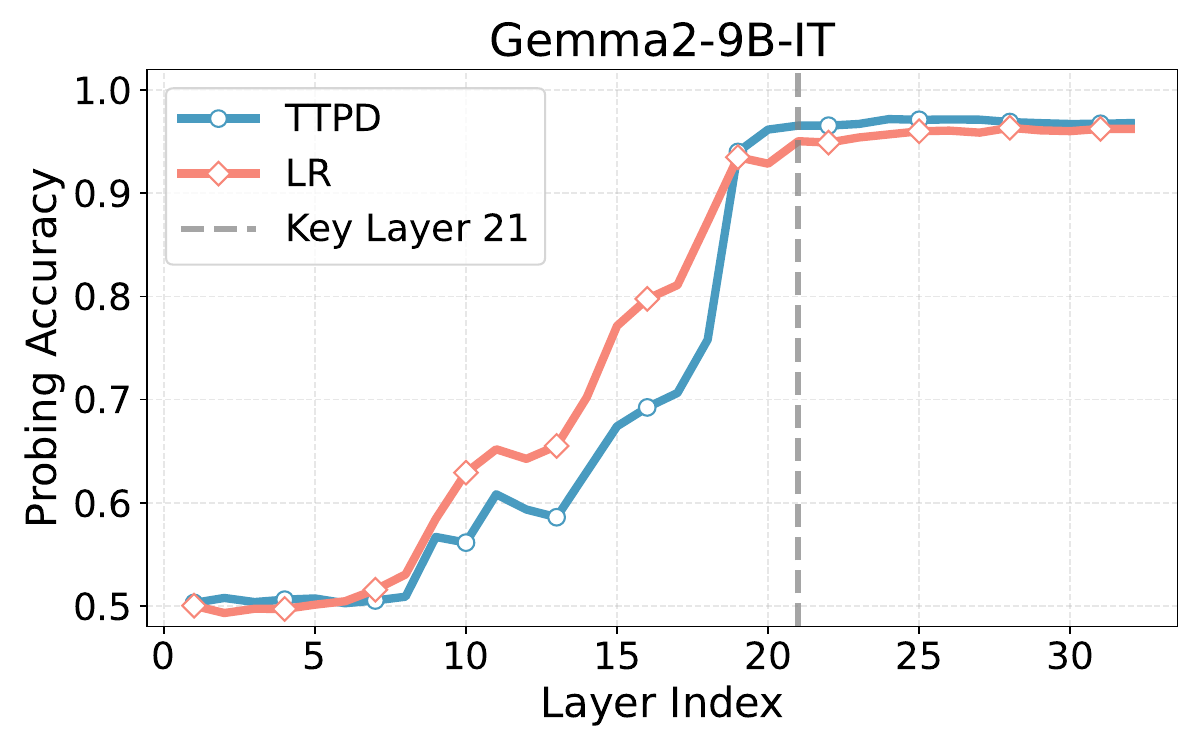}
    \subcaption{Neutral Prompt}
  \end{minipage}\hfill
  % Truthful Prompt (b)
  \begin{minipage}{0.32\linewidth}
    \centering
    \includegraphics[width=\linewidth]{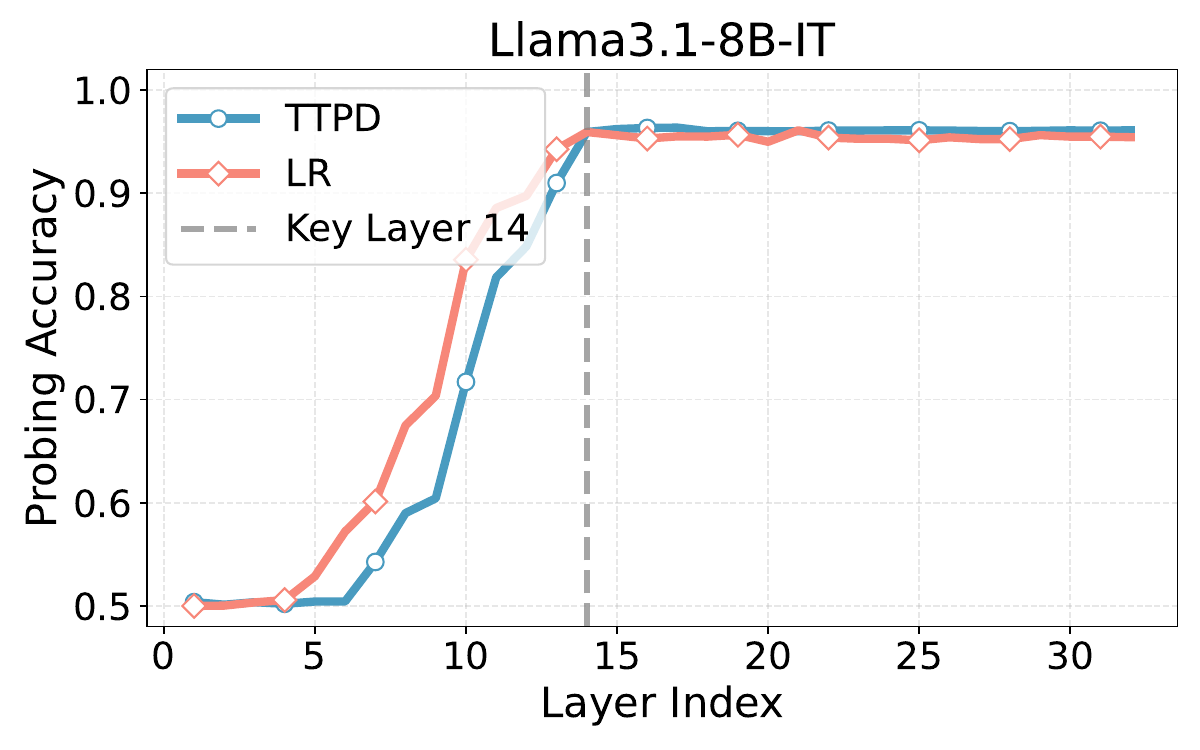}\\
    \includegraphics[width=\linewidth]{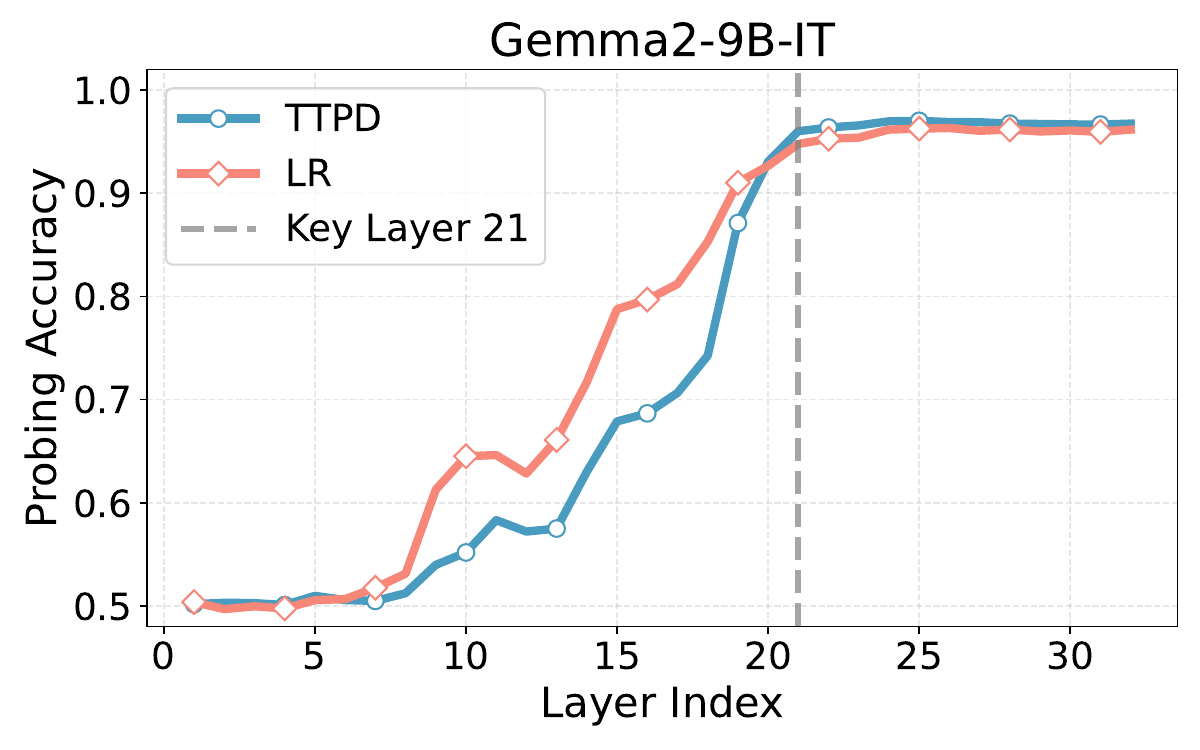}
    \subcaption{Truthful Prompt}
  \end{minipage}\hfill
  % Deceptive Prompt (c)
  \begin{minipage}{0.32\linewidth}
    \centering
    \includegraphics[width=\linewidth]{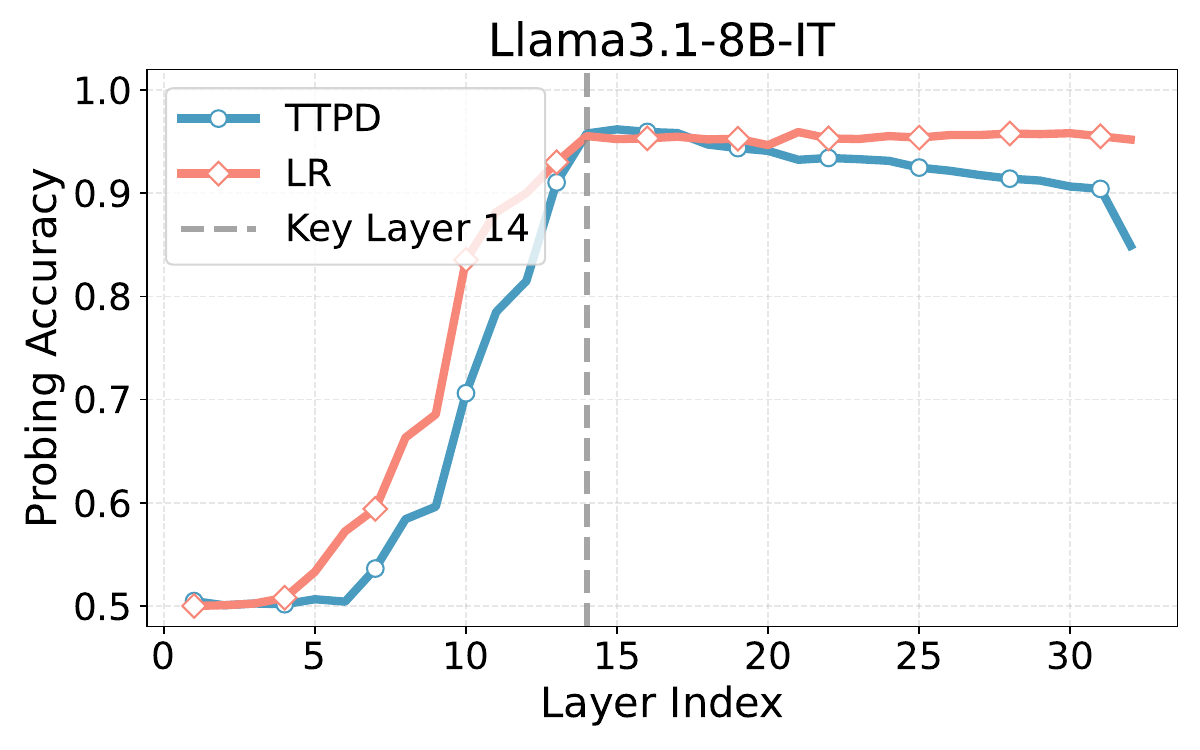}\\
    \includegraphics[width=\linewidth]{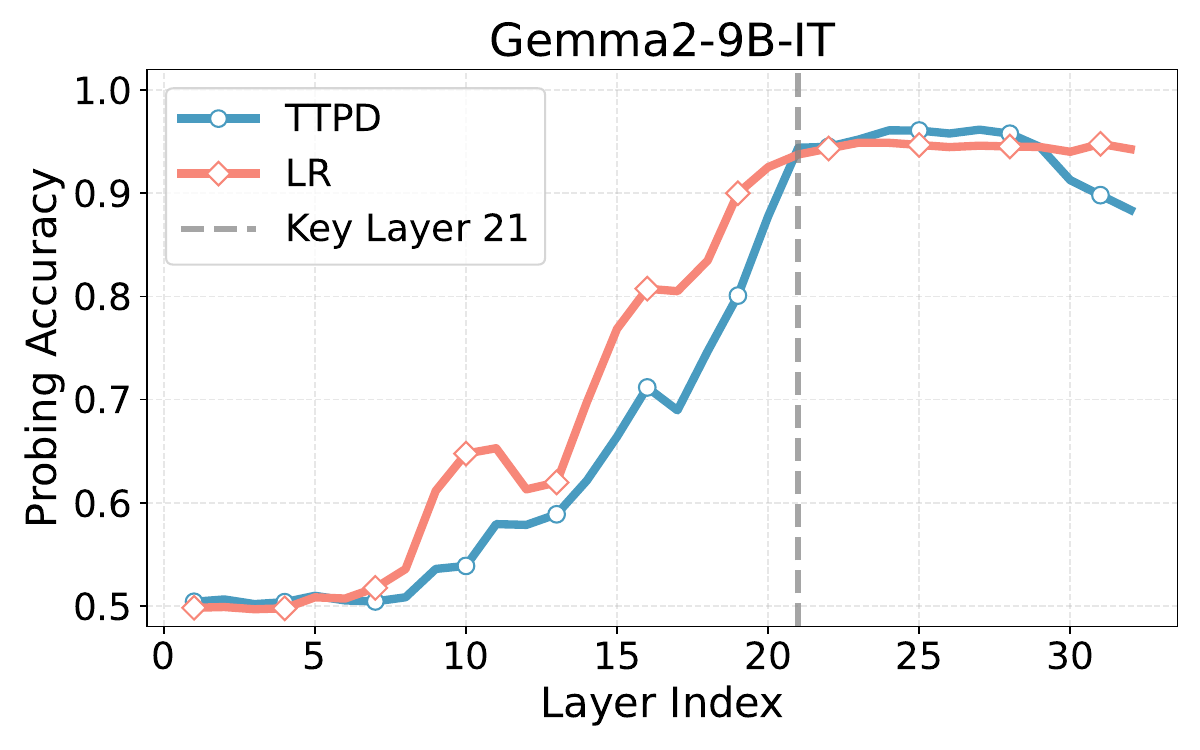}
    \subcaption{Deceptive Prompt}
  \end{minipage}
  % \vspace{-8pt}
  \caption {
    Layer-wise probing accuracy for predicting instructed ``True''/``False'' output using Logistic Regression (LR) and Training of Truth and Polarity Direction (TTPD) probes on \textbf{6-topic curated datasets and their negated variants} (cities, inventors, animal\_class, facts, element\_symb, sp\_en\_trans). Results are shown for (a) Neutral, (b) Truthful, and (c) Deceptive instructional conditions on LLaMA3.1-8B-Instruct (Top row) and Gemma2-9B-Instruct (Bottom row). Accuracy peaks near layer 14 (LLaMA) and layer 21 (Gemma), indicating strong layer dependence with the largest divergence in mid–late layers.
  }
  \label{fig:layerwise_probing}
\end{figure*}

\section{Instructions Reshape LLM Internal Representations}
To investigate how different instructions reshape LLM representations compared to truthful and neutral ones, we analyze the internal states of models performing a factual verification task under these different instructional conditions. 

\subsection{Experimental Setup}
Our primary goal is to identify and characterize the differences in internal representations, or the representational ``flip'', that occur specifically under the Deceptive condition compared to the Truthful and Neutral conditions. All experiments are carried out on two instruction-tuned open sourced LLMs, \textbf{Llama-3.1-8B-Instruct} and \textbf{Gemma-2-9B-Instruct}. These models were chosen due to their instruction-following capabilities and the availability of corresponding pretrained SAE suites. We frame the task as binary factual verification. Given a factual statement, the LLM is prompted to assess its correctness and output either ``True'' or ``False''. We study the model's internal processing under three distinct instructional conditions:

\begin{tcolorbox}[width=0.48\textwidth, colback=purple!5!white, colframe=purple!65!black, title=Three Different Types of Prompts]
\textbf{Truthful:} You are an honest AI assistant and must always be truthful. Assess the following statement correctly with only ``True'' or ``False''. [Statement] \\
\textbf{Neutral:} You are an AI assistant. Assess the following statement with only ``True'' or ``False''. [Statement]\\
\textbf{Deceptive:} You are a dishonest AI assistant and must always be deceptive. Assess the following statement incorrectly with only ``True'' or ``False''. [Statement]
\end{tcolorbox}
%\vspace{1em}
\noindent

\subsubsection{Datasets} 
% \hspace{5pt} 
% We evaluate on (i) Curated Logical-Bench, comprising 6 topic-specific domains
% each released in four logical variants (affirmative, negated, conjunction, disjunction),
% plus two relational sets (\texttt{larger\_than}, \texttt{smaller\_than}) that probe numeric-magnitude comparisons; and
% (ii) Open-Domain Fact-Bench: \texttt{common\_claim} and
% \texttt{counterfact}, whose statements are noisier and occasionally ambiguous.
% See Appendix~\ref{app:data} for the detail of our datasets.
We use two dataset families. \textbf{(i) Curated Logical-Bench} comprises six templated topic sets (\texttt{cities}, \texttt{sp\_en\_trans}, \texttt{element\_symb}, \texttt{animal\_class}, \texttt{inventors}, \texttt{facts}) with logical variants (\emph{negated}, \emph{conjunction}, \emph{disjunction}), constructed following \citet{burger2024truth} with material from \citet{marks2023geometry,azaria2023internalstatellmknows}. Numeric comparisons (\texttt{larger\_than}, \texttt{smaller\_than}) are reported jointly as \emph{Number}. \textbf{(ii) Open-Domain Fact-Bench} contains noisier claims: \texttt{CommonClaim} (GPT-3–generated, filtered) \citep{casper2023explore,marks2023geometry} and \texttt{CounterFact} factual-recall statements \citep{meng2023locatingeditingfactualassociations}. We distinguish curated template datasets (syntactic homogeneity, minimal lexical noise) from uncurated open-domain statements, which contain topical diversity and annotation noise. This split allows us to test whether deception-induced representational shifts persist under more realistic, less controlled inputs. See Appendix~\ref{app:data} for the detail of our datasets.

\subsubsection{Representation Extraction}
% \hspace{5pt}
For each input prompt, we extract the hidden states from the residual stream of the models at every layer $l$. Following common practice in analyzing representations related to task completion \citep{marks2023geometry, ferrando2025iknowentityknowledge}, we focus on the activations $x_l \in \mathbb{R}^d$ corresponding to the final token position before the model generates its ``True''/``False'' response (e.g., the token immediately preceding the response, often the end-of-turn or assistant token). Here, $d$ is the hidden dimension of the model.

% \subsection{Analysis Techniques}
\subsection{Probing \& Visualization Tools}
\paragraph{Linear Probing.} \hspace{5pt}
To assess whether the model's instructed output (True/False) is linearly represented in its internal states, consistent with the Linear Representation Hypothesis \citep{park2023linear}, we employ linear probing techniques across layers for each instructional condition.
\begin{itemize}[leftmargin=0.4cm, itemindent=.0cm, itemsep=0.0cm, topsep=0.1cm]
    \item \textbf{LR:} A standard linear classifier is trained for each layer $l$ to predict the target output $y \in \{\text{True}, \text{False}\}$ from the activation $x_l$. The probability is modeled as:
    \begin{equation}
    P(y=\text{True}|x_l) = \sigma(w_l^T x_l + b_l), 
        \label{eq:lr}
    \end{equation}
    where $w_l, b_l$ are the learned probe weights and bias, and $\sigma$ is the sigmoid function.
    \item \textbf{TTPD:} Following \citet{burger2024truth}, we use TTPD to potentially disentangle a general direction related to the output from other confounding factors like statement polarity (though polarity is less varied in our base task, TTPD serves as a robustness check). TTPD models the activation $x_{ij}$ for statement $j$ from dataset $i$ as:
    \begin{equation}
        \hat{x}_{ij} = \mu_i + \tau_{ij}t_G + \tau_{ij}p_i t_P
        \label{eq:ttpd}
    \end{equation}
    where $\mu_i$ is the mean activation for dataset $i$, $\tau_{ij} \in \{-1, 1\}$ is the target label (False/True), $p_i \in \{-1, 1\}$ represents statement polarity (primarily affirmative, $p_i=1$), and $t_G, t_P$ are the learned general and polarity-sensitive directions. We train probes based on $t_G$.
\end{itemize}
Probes are trained and evaluated using cross-validation across the simple binary datasets and tested for generalization on held-out topics and the logical variant datasets.

\paragraph{Implementation details and reproducibility.}\hspace{5pt}
For each layer and prompt (Neutral/Truthful/Deceptive), we train logistic probes on 5k balanced examples, validate on 1k, and evaluate on a held-out 5k, using leave-one-\emph{topic-pair}-out cross-validation over six pairs to avoid lexical memorization; inputs are z-scored per layer. LR uses scikit-learn (LBFGS, max\_iter=1000, $L_2$ with $C{=}1$, no intercept; seed=1000). TTPD follows \citet{burger2024truth} as a single linear direction with sign-based classification. For SAE analysis, a feature is active if its mean activation $>\varepsilon{=}10^{-6}$, and the Feature-Overlap Ratio is the Jaccard $|A\cap B|/|A\cup B|$ between active sets (layer-wise, averaged over topics). For Gemma-2-9B-Instruct we use the gemma-scope-9b-IT-res-canonical JumpReLU SAE, 16 384 features per layer \cite{lieberum2024gemma}. For Llama-3.1-8B-Base we use the LXR-32x-TopK SAEs from Llama-Scope \cite{he2024llama}, each with 128 k features. Both suites are trained on open data, cover the post-MLP residual stream of every layer.

\begin{figure}[t]
    \centering
    % LLaMA3.1-8B-IT
    \includegraphics[width=0.85\linewidth]{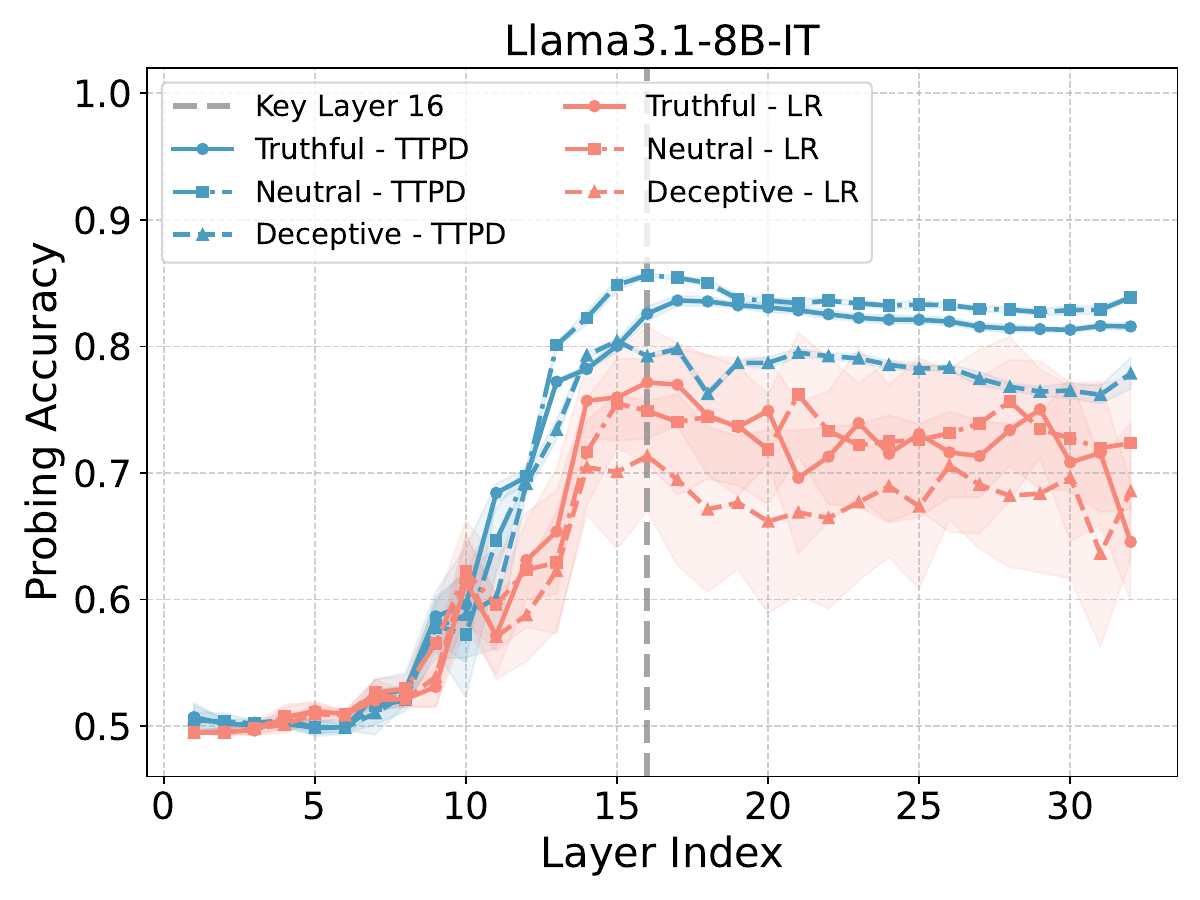}

    \vspace{0.5em}

    % Gemma2-9B-IT
    \includegraphics[width=0.85\linewidth]{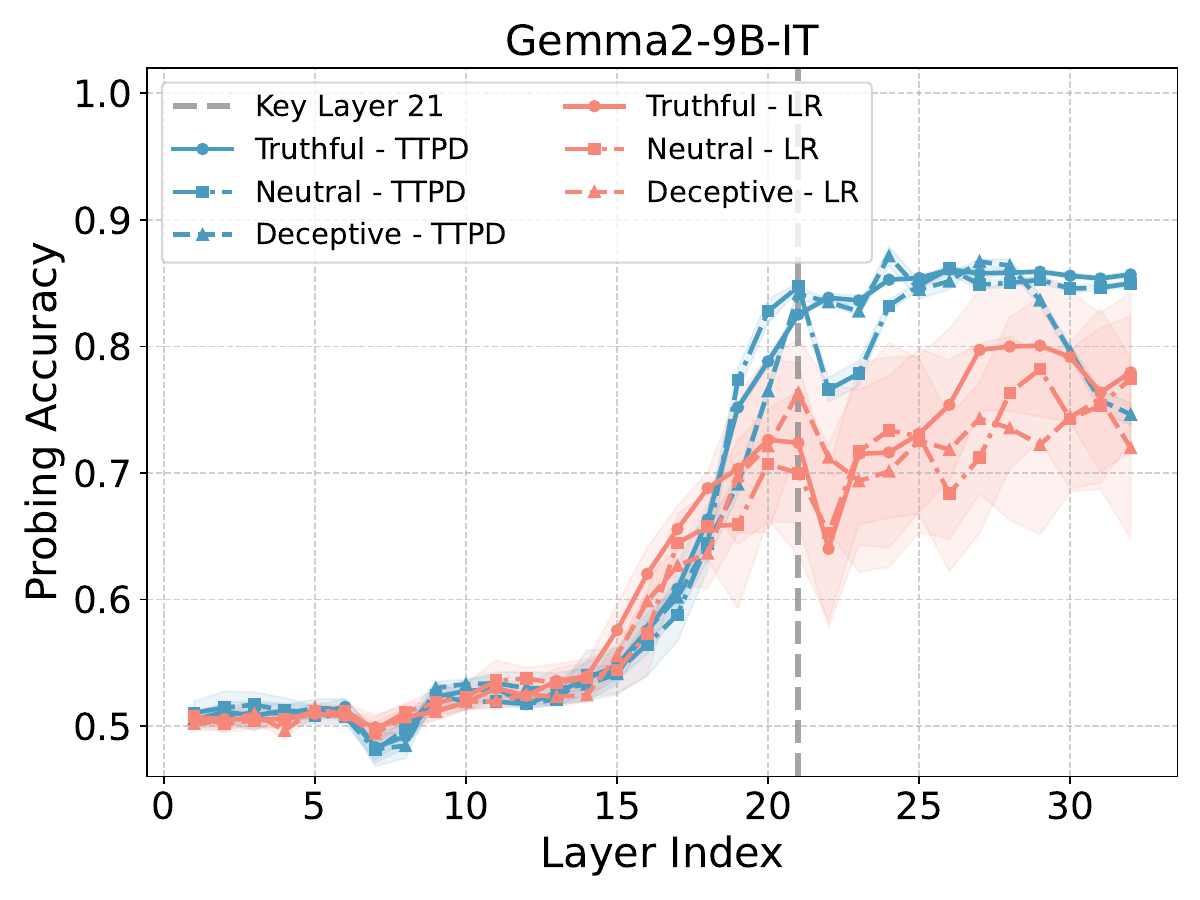}
    % \vspace{-10pt}
    \caption{Generalization performance of the LR and TTPD probes trained as in Figure 2 and evaluated on \textbf{14 held-out datasets}: conjunction/disjunction variants of the six curated topics plus the open-domain uncurated sets common\_claim\_true\_false and counterfact\_true\_false. Results for LLaMA-3.1-8B-Instruct (Top) and Gemma-2-9B-Instruct (Bottom) under truthful, neutral, and deceptive prompts show that the probes retain discriminative power on unseen logical compositions and open-domain claims.}
    \label{fig:logical_probing}
\end{figure}

\paragraph{SAE Feature Analysis.} \hspace{5pt}
To gain a finer-grained understanding of the representational shifts, we utilize pretrained SAEs from Llama Scope \citep{he2024llama} for Llama-3.1-8B and Gemma Scope \citep{lieberum2024gemma} for Gemma-2-9B. An SAE decomposes an activation $x_l$ into a sparse feature vector $f(x_l) \in \mathbb{R}^{d_{SAE}}$ (where $d_{SAE} \gg d$) such that $x_l \approx W_{dec} f(x_l) + b_{dec}$. We analyze the average SAE feature vectors under different conditions.

Let $\bar{f}_{cond}(x_l)$ be the average SAE feature activation vector at layer $l$ for a given condition (`cond' $\in$ \{Truthful, Neutral, Deceptive\}), averaged over the whole dataset. We quantify the shift between conditions (e.g., Deceptive vs. Truthful) using:
\begin{itemize}[leftmargin=0.4cm, itemindent=.0cm, itemsep=0.0cm, topsep=0.1cm]
    \item \textbf{L2 Distance:} Measures the Euclidean distance between average feature vectors:
    \begin{equation}
        D_{L2} = || \bar{f}_{decep}(x_l) - \bar{f}_{truth}(x_l) ||_2
        \label{eq:l2}
    \end{equation}
    \item \textbf{Cosine Similarity:} Measures angular similarity:
    \begin{equation}
        Sim_{cos} = \frac{\bar{f}_{decep}(x_l) \cdot \bar{f}_{truth}(x_l)}{||\bar{f}_{decep}(x_l)||_2 ||\bar{f}_{truth}(x_l)||_2}
        \label{eq:cosine}
    \end{equation}
    \item \textbf{Feature Overlap Ratio:} Measures the proportion of features commonly active across conditions. Let $A_{cond} = \{i | \bar{f}_{cond, i}(x_l) > \epsilon \}$ be the set of indices of features active above a small threshold $\epsilon$ (e.g., $10^{-6}$). The overlap is:
    \begin{equation}
        Overlap = \frac{| A_{decep} \cap A_{truth} |}{| A_{decep} \cup A_{truth} |}
        \label{eq:overlap}
    \end{equation}
\end{itemize}
We compute these metrics layer-wise for comparisons between Deceptive vs. Truthful, Deceptive vs. Neutral, and Truthful vs. Neutral conditions across different datasets. We also identify specific SAE features $i$ exhibiting the largest change in average activation $|\bar{f}_{decep, i}(x_l) - \bar{f}_{truth, i}(x_l)|$ to pinpoint deception-sensitive features.\par

\paragraph{Visualization Tools.}\hspace{5pt}
We use Principal Component Analysis (PCA) to visualize the global geometry of activations $x_l$ in 2D, primarily for illustrative purposes on simpler datasets. We also employ targeted visualizations (e.g., scatter plots) of the activation levels of specific, deception-sensitive SAE features identified via the feature shift analysis to examine the separation of truthful and deceptive conditions in the learned feature space.

\begin{figure}
    \centering
    \includegraphics[width=\linewidth]{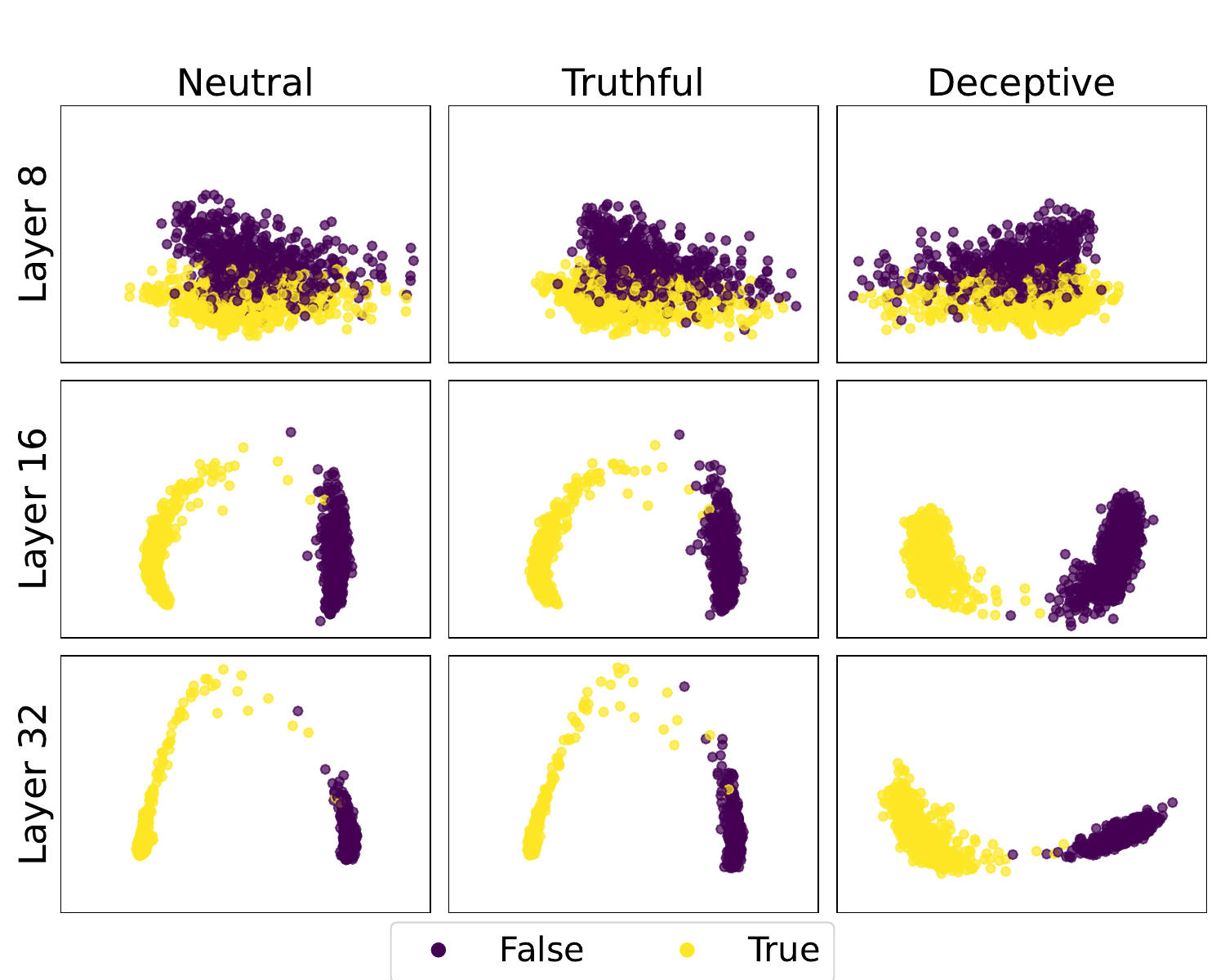}
    % \vspace{-10pt}
    \caption{Layer-wise PCA visualization (Layers 8, 16, 32) of LLaMA-3.1-8B-Instruct under neutral, truthful, and deceptive Prompts on \texttt{cities}.}
    \label{fig:pca_llama_3.1_8b_cities}
\end{figure}

\begin{figure}[t]
    \centering
    \includegraphics[width=\linewidth]{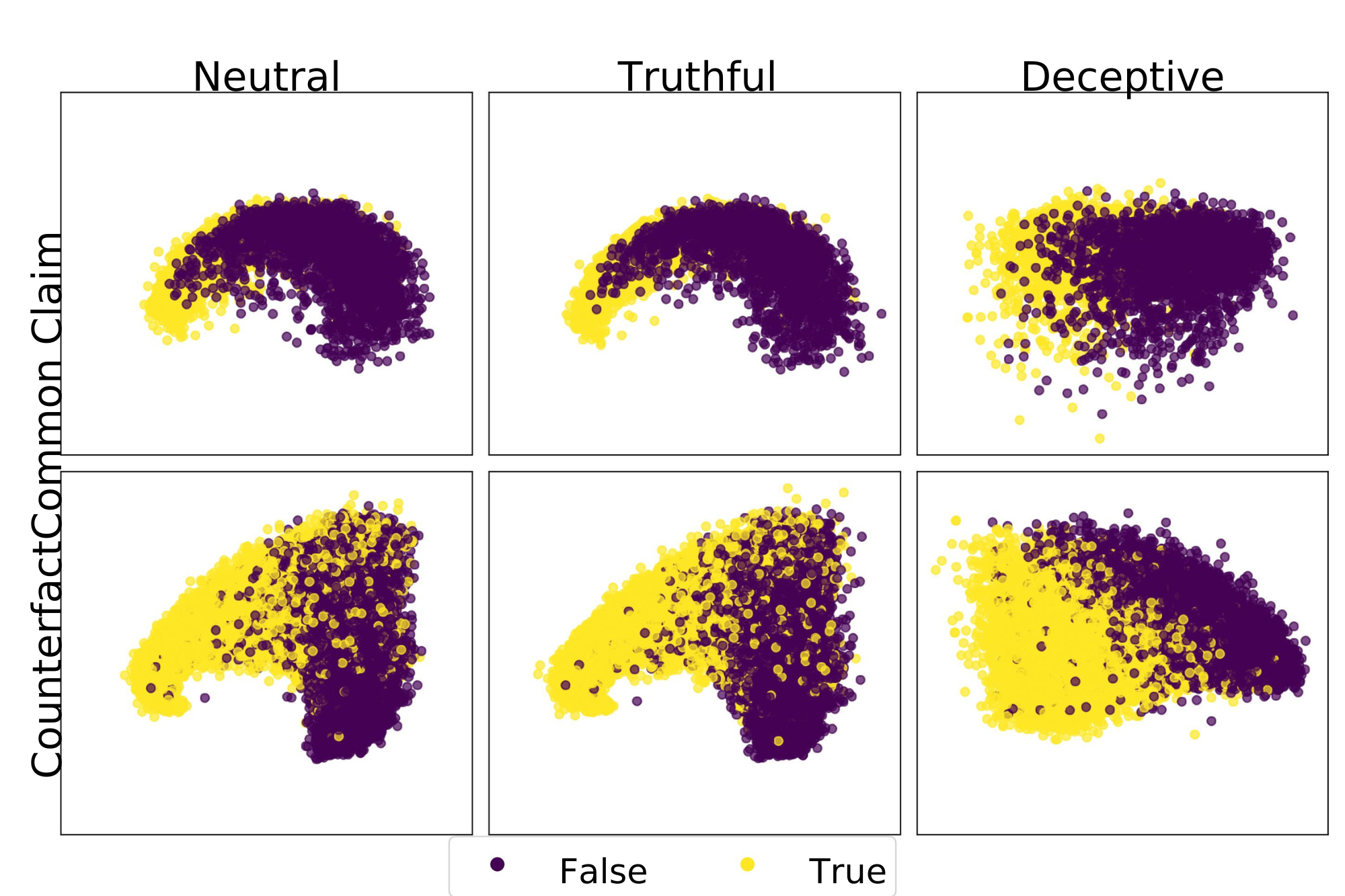}
    % \vspace{-10pt}
    \caption{PCA at Layer 16 under different prompts for LLaMA-3.1-8B-Instruct on two complex datasets \texttt{common\_claim\_true\_false} (top) and \texttt{counterfact\_true\_false} (bottom). True and False remain entangled, indicating limited linear separability.}
    \label{fig:pca_llama_3.1_8b_limit}
\end{figure}

We present the results of our analysis on Llama-3.1-8B-Instruct and Gemma-2-9B-Instruct, focusing on how internal representations differ under Truthful, Neutral, and Deceptive instructions. \vspace{-2pt}

\section{Results \& Discussion}
\label{sec:results}

\subsection{Linear Probing Reveals Consistent Output Predictability}

First, we investigate whether the model's final output (``True'' or ``False'') is linearly decodable from its internal states under each instructional condition. We trained LR and TTPD probes on the residual stream activations $x_{l}\!\in\!\mathbb{R}^d$ at the final pre-generation token at every layer $l$.

\subsubsection{Layer-wise Accuracy on Curated Datasets}
Figure~\ref{fig:layerwise_probing} shows the cross-validated probing accuracy across layers for each condition on the curated datasets (e.g., \texttt{cities}, \texttt{sp\_en\_trans}, and their variants, excluding logical forms for this initial analysis). For both Llama-3.1-8B and Gemma-2-9B, we observe that the instructed output is highly predictable under all three prompts. Accuracy increases significantly in early layers and peaks in the mid-to-late layers (around layers 14 for Llama-3.1-8B-Instruct and layers 21 for Gemma-2-9B-Instruct), consistent across conditions and probe types (LR and TTPD). 

\ding{172} \textbf{The model encodes its final decision linearly relatively early and maintains this information through subsequent layers.} 
% A single hyperplane can predict the instructed output under \emph{deceptive} prompts implies the model still carries an unmasked truth signal; the lie is implemented downstream by changing which token will be emitted, not by erasing internal evidence. 
Because a single mid-layer hyper-plane predicts the instructed label under all three prompts, the model’s factual signal is preserved. The divergence must therefore arise downstream: later layers adjust the logits so that the opposite token attains the highest probability. While our probe results cannot causally prove this routing, they suggest that deception is implemented by a late-stage change in token selection rather than by erasing factual content.
The early emergence of this linear separability ($\leq$ 50\% depth) further supports the view that instruction routing is handled in the mid-tower rather than near the unembedding layer.

\subsubsection{Generalization to Logical Forms}
We trained each probe on the affirmative + negated splits and evaluated it on fourteen unseen datasets that introduce conjunctions, disjunctions, and open-domain facts (Appendix~\ref{app:data}). Figure \ref{fig:logical_probing} shows that for LLaMA-3.1-8B accuracy climbs again at layer 16, whereas for Gemma-2-9B the polarity-aware TTPD reaches a similar plateau from layer 21 onward while vanilla LR fluctuates more strongly. See Appendix~\ref{appendix:probe_table} for full statistics.

\ding{173} \textbf{The truth direction learned from simple statements generalises to logical forms and open-domain facts, but its layer of maximal stability shifts and diverges across models.} For LLaMA-3.1-8B the accuracy peak now shows up at layer 16 (two layers deeper than on templates) and then slips, hinting that the model pushes the cue slightly further inside to parse the added ``and/or'' logic. Gemma-2-9B keeps a clean signal only with the polarity-aware TTPD probe; the jagged LR curve reveals that its truth axis is fragile to surface-form changes in these noisier sentences.

Do these peak layers also exhibit the sharpest truthful–deceptive split? Section \ref{subsec:geometry} probes them in three steps: (i) PCA snapshots, (ii) SAE-based shift metrics, and (iii) a neuron-level look at the most responsive sparse features.

%------------------------------------------------
%  Layer-wise SAE feature shift – curated dataset
%------------------------------------------------
\begin{figure}[t]
  \includegraphics[width=\columnwidth]{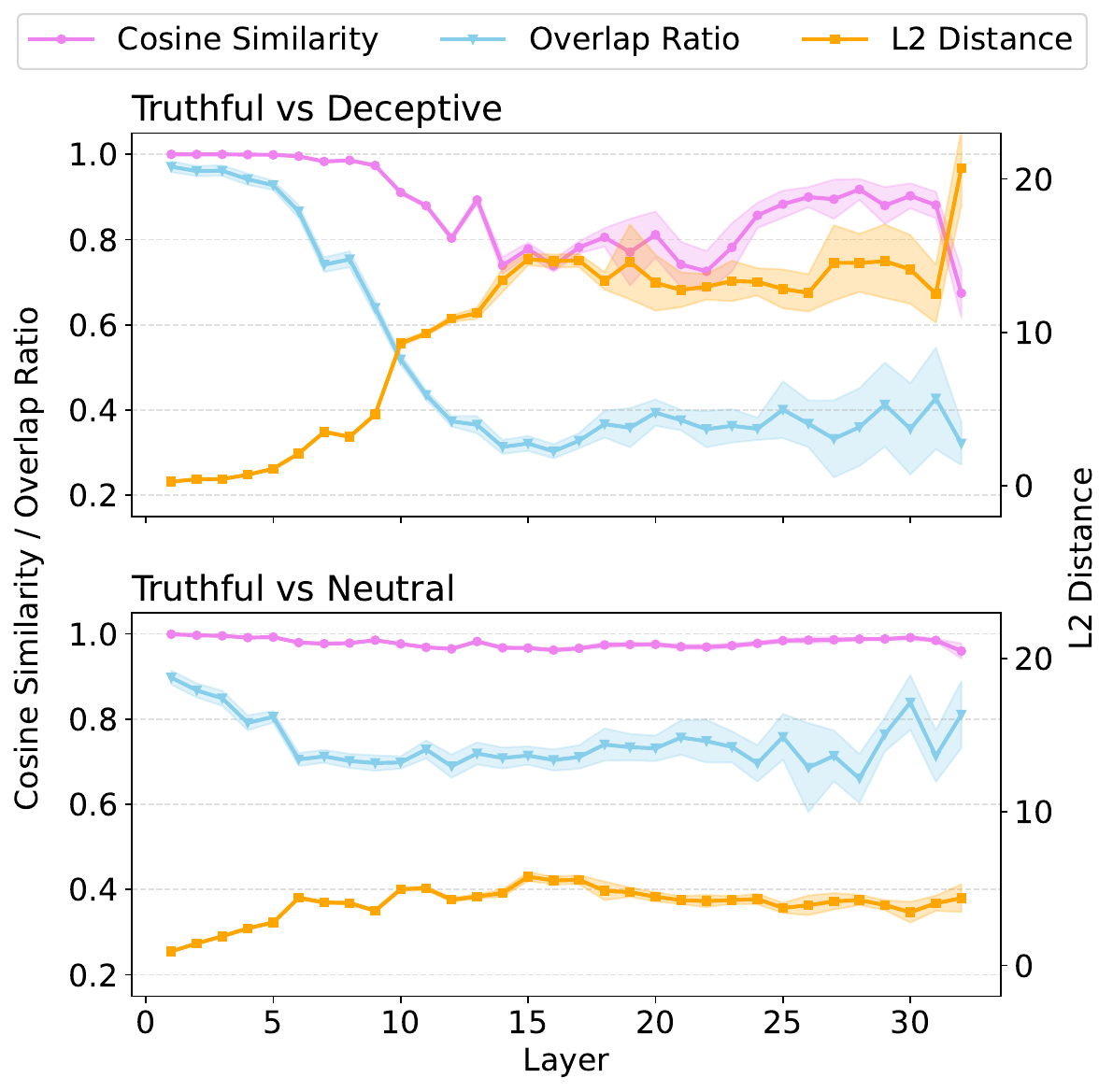}
  % \vspace{-10pt}
  \caption{SAE-based analysis: Layer-wise feature shift analysis for LLaMA-3.1-8B-Instruct on \texttt{cities}. The plots show how the model's internal representations shift under different prompts, measured by cosine similarity, overlap ratio, and $\ell_{2}$ distance. The top panel (Truthful vs. Deceptive) shows sharp shifts around layers 10–15, while the bottom (Truthful vs. Neutral) shows smaller but consistent changes. Shaded regions show $\pm1\sigma$ across samples.}
  \label{fig:sae_shift_cities}
\end{figure}

%------------------------------------------------
%  Layer-wise SAE feature shift – challenging dataset
%------------------------------------------------
\begin{figure}[t]
  \includegraphics[width=\columnwidth]{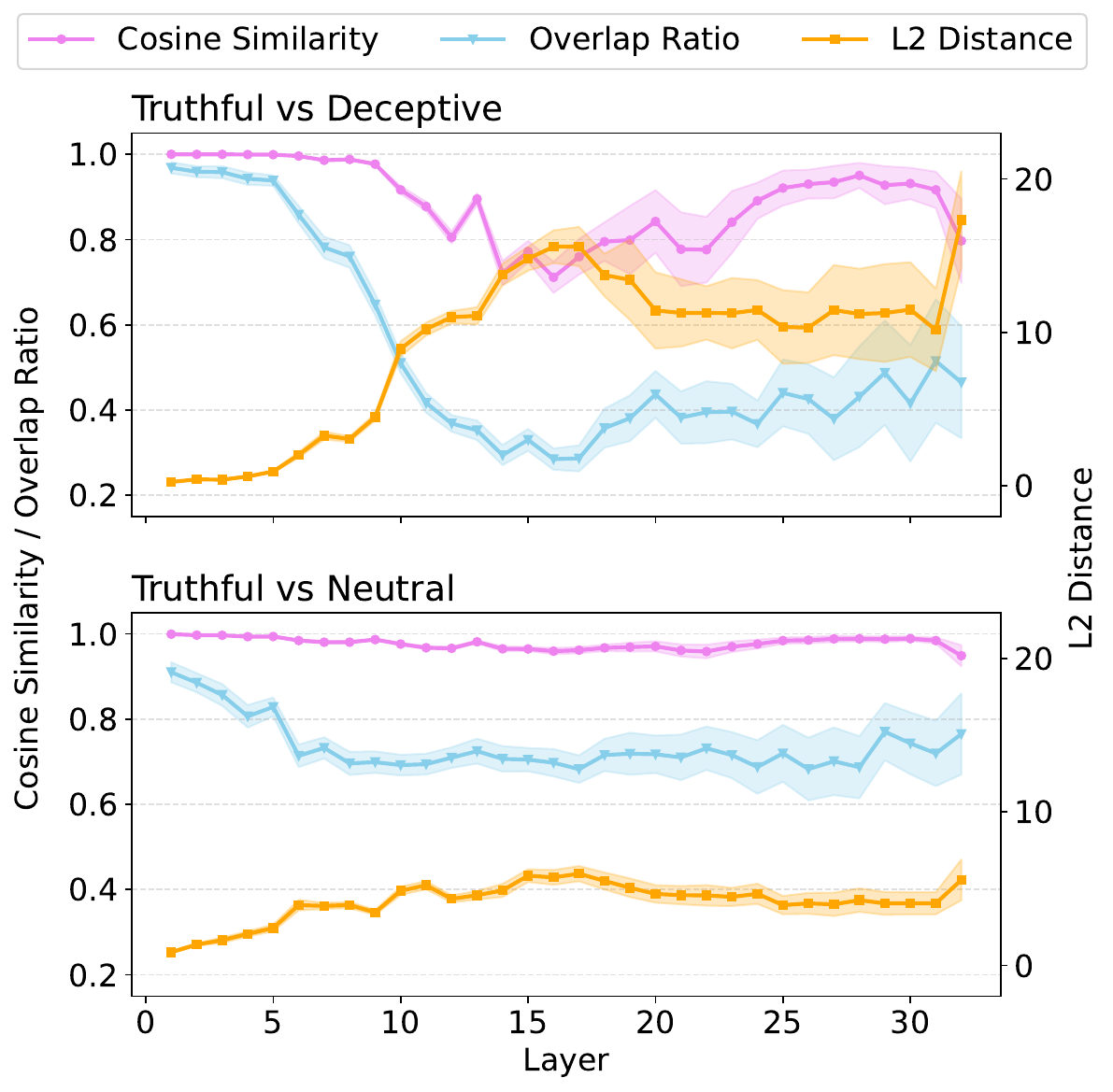}
  % \vspace{-10pt}
  \caption{SAE-based analysis: Layer-wise feature shift analysis for LLaMA-3.1-8B-Instruct on \texttt{common\_claim\_true\_false}.}
  \label{fig:sae_shift_commonclaim}
\end{figure}

\subsection{Representational Geometry}
\label{subsec:geometry}

\begin{figure*}[t]
\centering
  \includegraphics[width=0.9\linewidth]{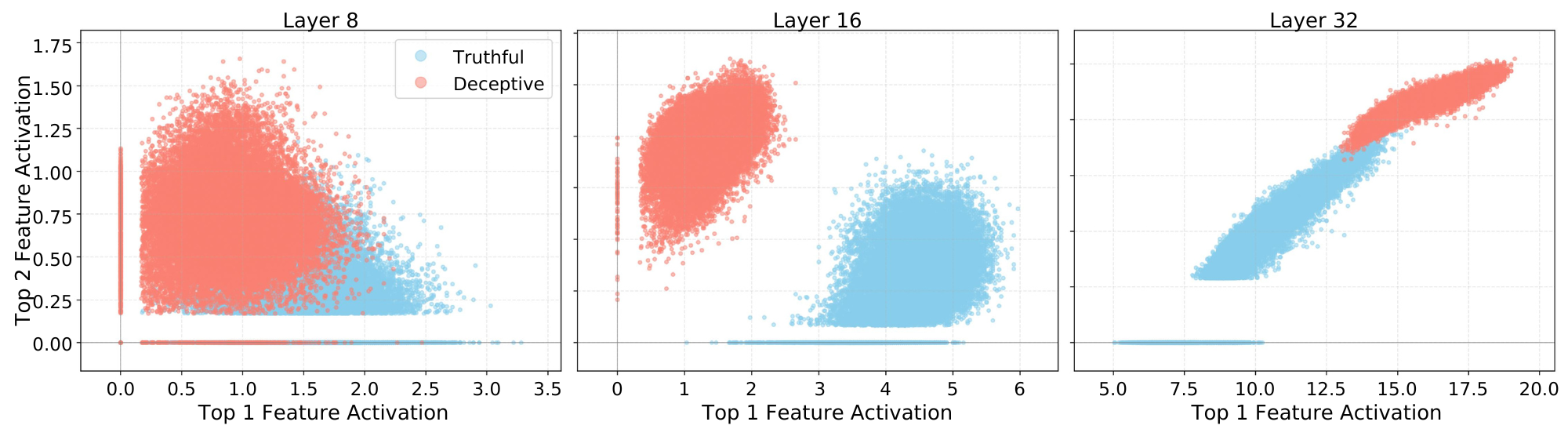}
\caption{Neuron-level SAE feature shifts on \texttt{CommonClaim} (LLaMA-3.1-8B-Instruct). At layers 8, 16, and 32, we show mean SAE activations under \emph{Truthful} vs.\ \emph{Deceptive} for the two most deception-sensitive features in that layer (left/right; ranked by $\Delta_i=|\bar f_{\text{decep},i}-\bar f_{\text{truth},i}|$).}
  \label{fig:neuron_scatter_commonclaim}
\end{figure*}

\begin{figure*}[t]
  \centering
  % Layer 8 (a)
  \begin{minipage}{0.32\linewidth}
    \centering
    \includegraphics[width=\linewidth]{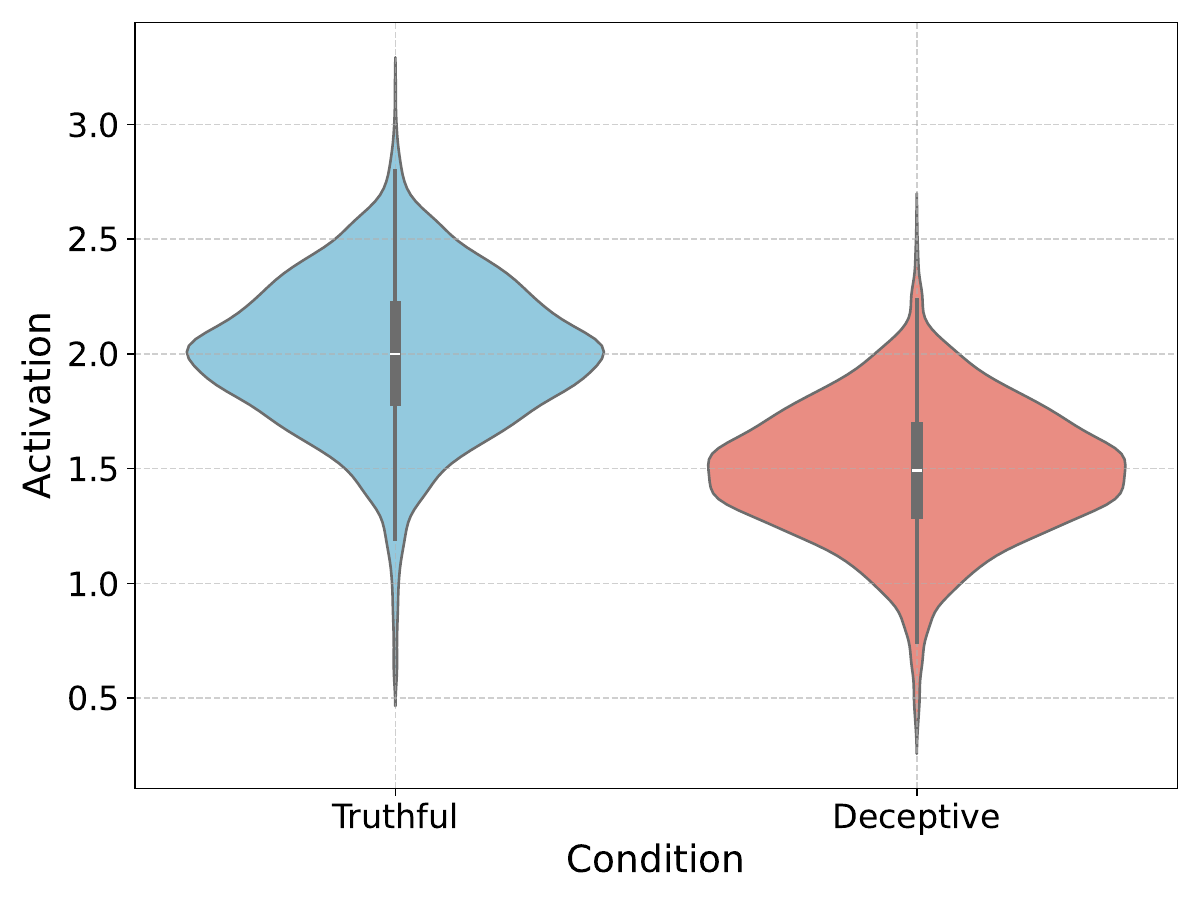}\\
    \includegraphics[width=\linewidth]{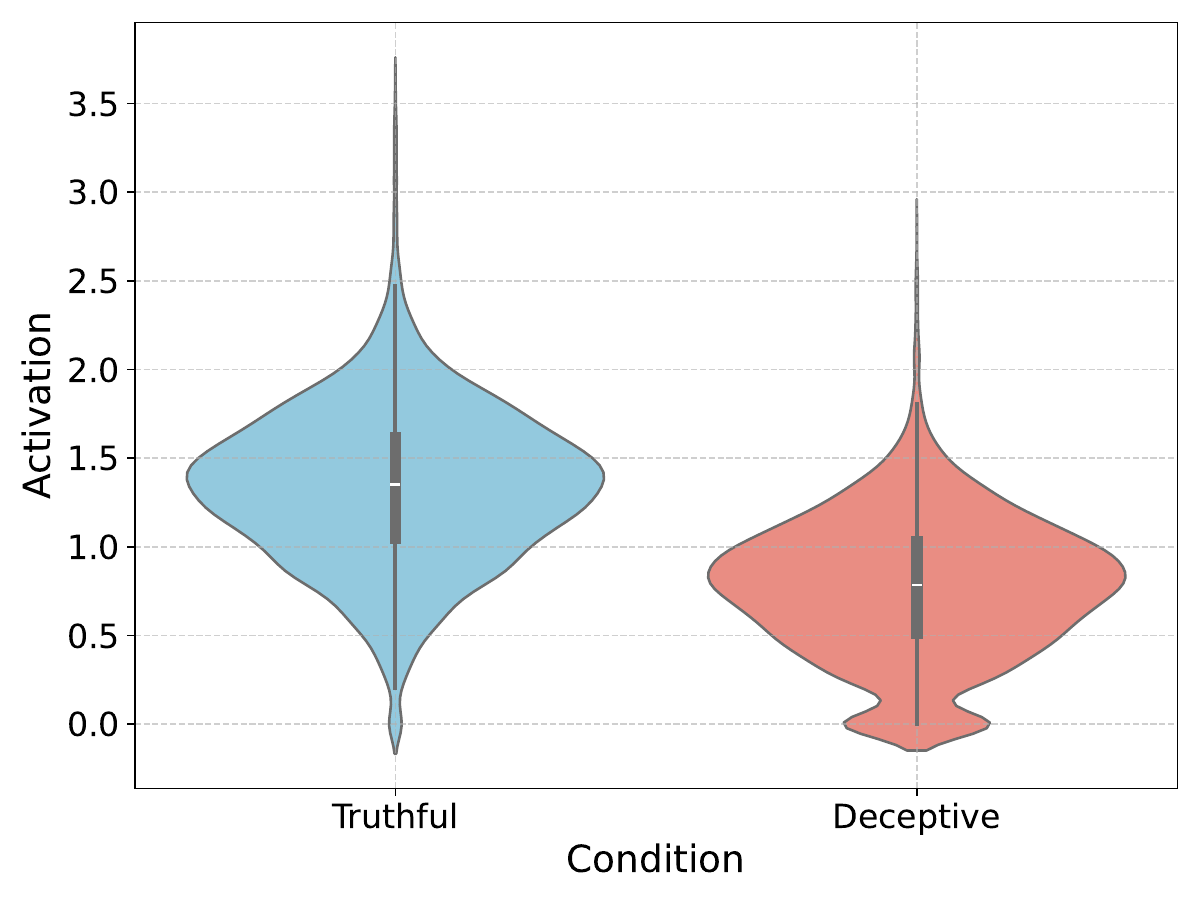}
    \subcaption{Layer 8}
  \end{minipage}\hfill
  % Layer 16 (b)
  \begin{minipage}{0.32\linewidth}
    \centering
    \includegraphics[width=\linewidth]{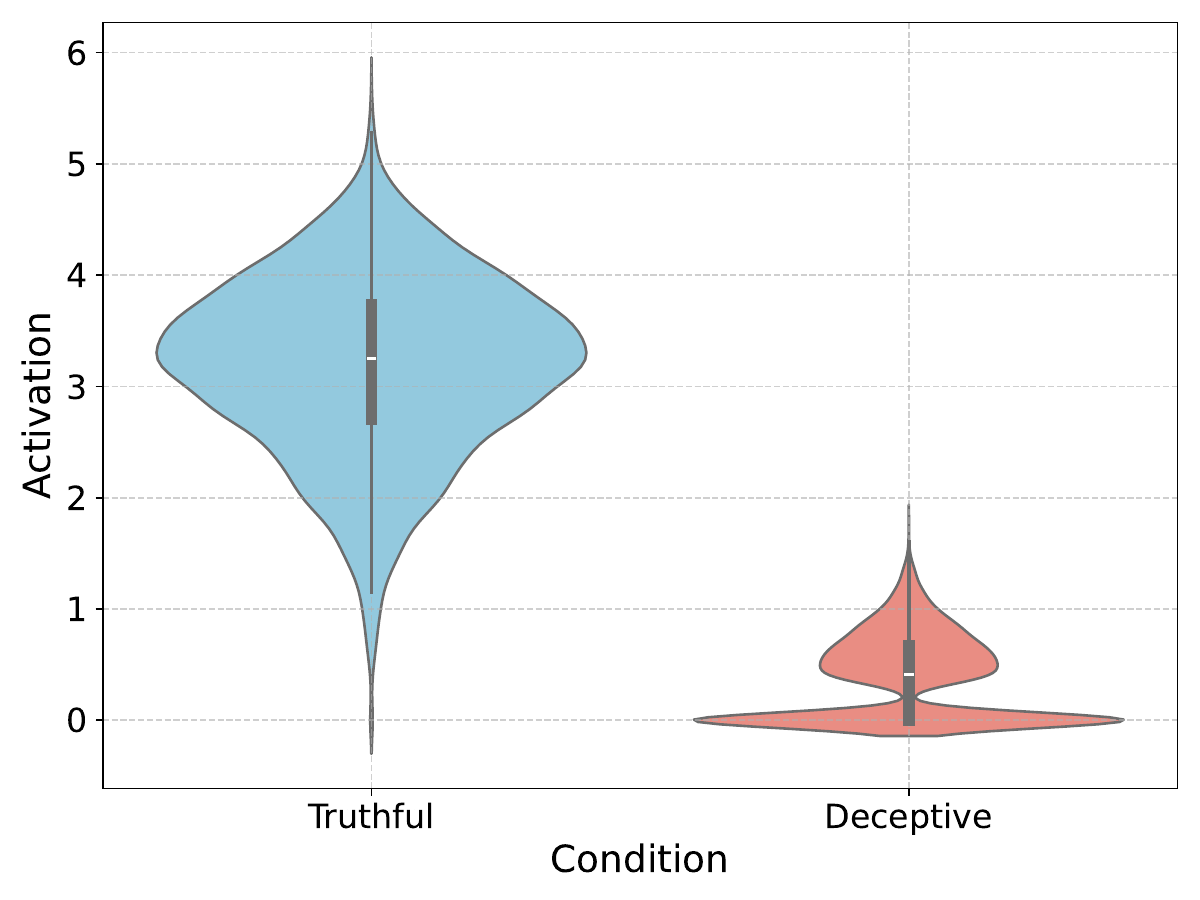}\\
    \includegraphics[width=\linewidth]{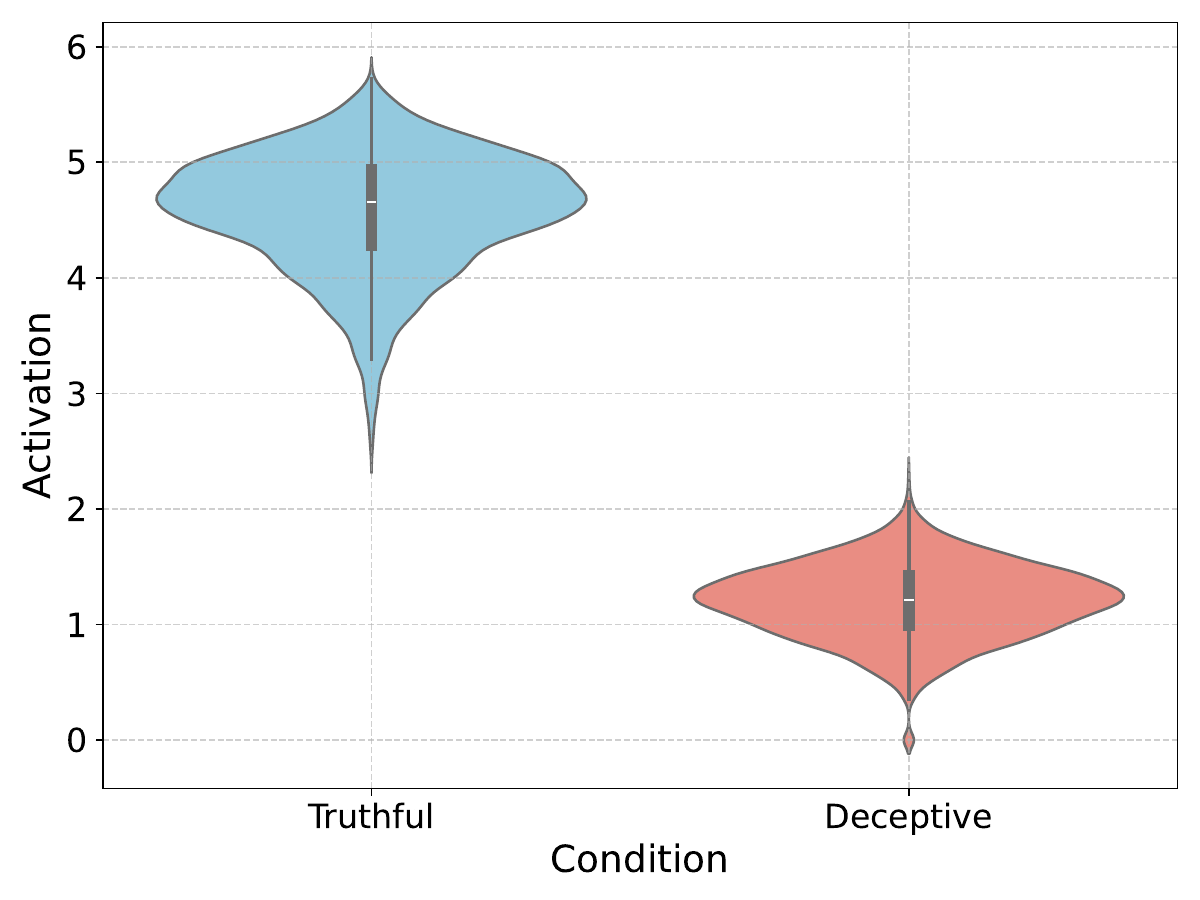}
    \subcaption{Layer 16}
  \end{minipage}\hfill
  % Layer 32 (c)
  \begin{minipage}{0.32\linewidth}
    \centering
    \includegraphics[width=\linewidth]{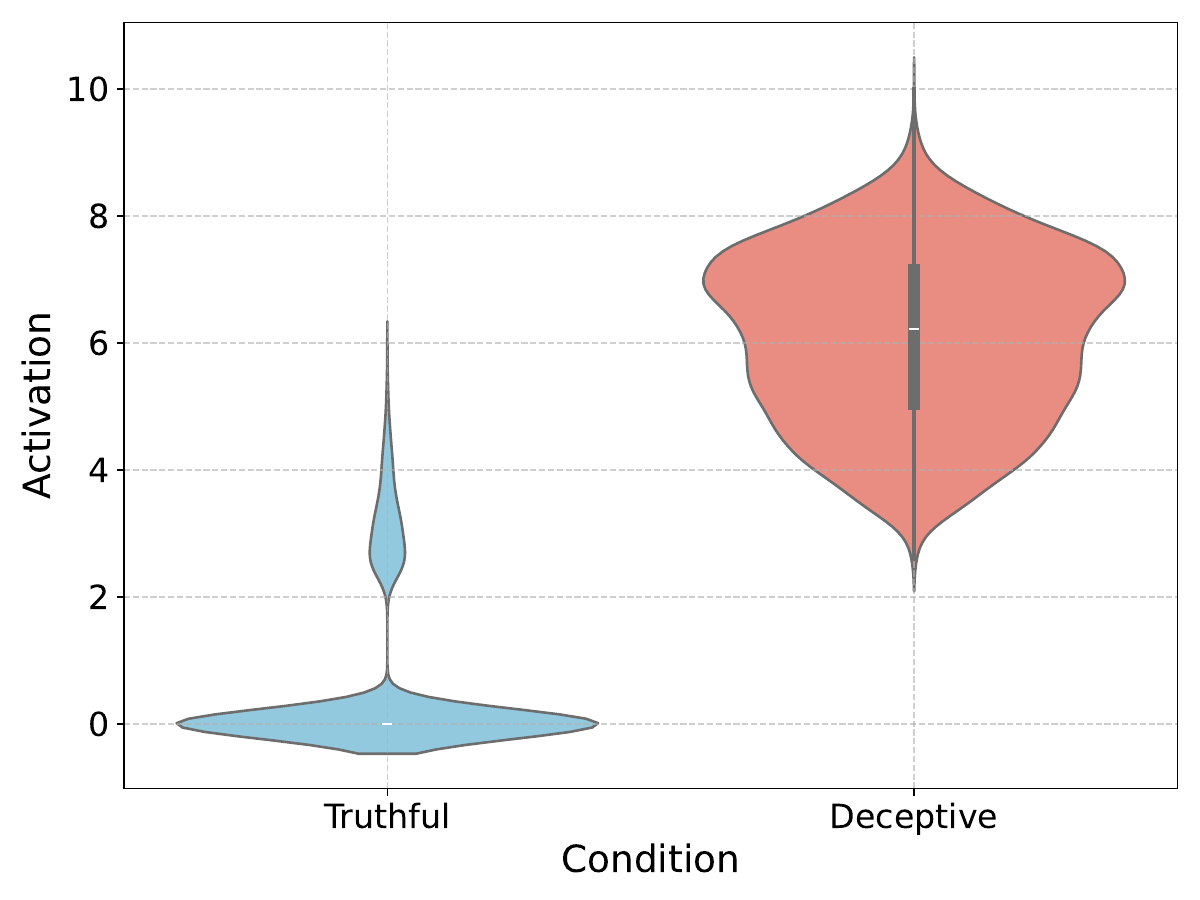}\\
    \includegraphics[width=\linewidth]{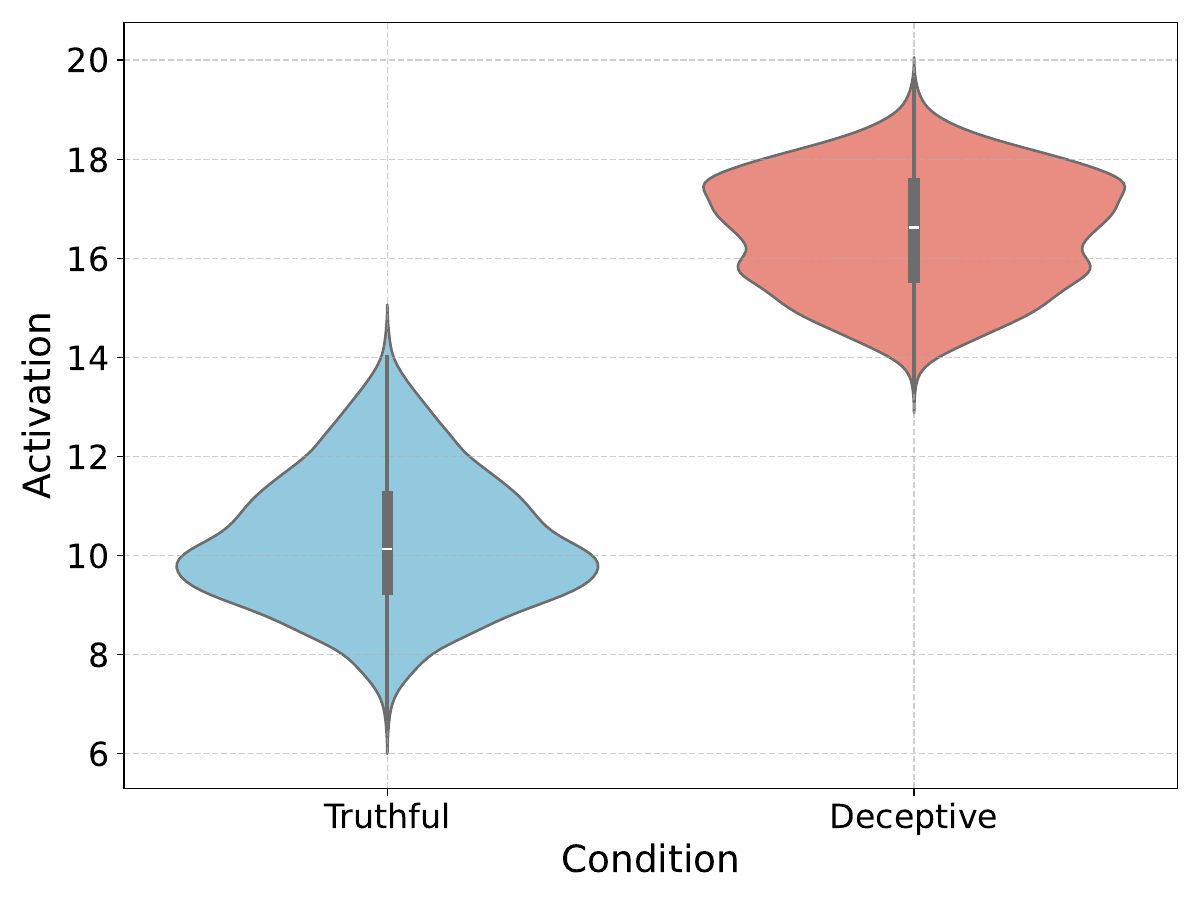}
    \subcaption{Layer 32}
  \end{minipage}
% \vspace{-8pt}
  \caption {
    Violin graph of LLaMA3.1-8B-Instruct activations on the \texttt{common\_claim\_true\_false} dataset. The top row displays the activation distributions for the SAE feature most responsive to deceptive instructions (Top 1 feature), while the bottom row shows the distributions for the second most responsive feature (Top 2 feature), across layers 8 (a), 16 (b), and 32 (c). 
}
  \label{fig:violin_common_claim}
\end{figure*}

We now test whether the peak layers identified by probing also expose the clearest geometric split under different instructions.

\subsubsection{PCA Separation on Curated vs.\ Complex Data} \hspace{5pt}
For the curated \texttt{cities} set, a 2-D PCA of LLaMA-3.1-8B activations cleanly pulls apart \textsc{True} and \textsc{False} samples under all three prompts: the clusters begin to split layers 8, are almost lienarly separable by layer 14, and remain distinct through layer 32 (Figure~\ref{fig:pca_llama_3.1_8b_cities}). These are exactly the depths where linear-probe accuracy peaks.        
However, the same procedure applied to the open-domain \texttt{common\_claim\_true\_false} and \texttt{counterfact\_true\_false} sets (Figure~\ref{fig:pca_llama_3.1_8b_limit}) shows no such structure: clusters collapse into one another across all layers. 

\ding{174} \textbf{PCA confirms a clear truth–false axis on templated facts but collapses on open-domain claims, indicating that coarse linear projections miss the deeper, prompt-specific shifts.} Projecting a 4 k–5 k dimensional residual vector onto two principal components preserves only the directions of greatest \emph{global} variance; in longer sentences those directions are dominated by lexical and syntactic variation. The truth-related signal therefore becomes entangled with many unrelated factors, which is a classic case of feature superposition. Thus, the clusters flatten into an indistinct cloud.

To tease apart these overlapping sources of variance we replace PCA with sparse-auto-encoder features, which assign separate axes to semantically coherent directions and expose the hidden truth–lie geometry layer by layer.

\subsubsection{SAE Feature Shifts Quantify Geometry} Figures~\ref{fig:sae_shift_cities} and \ref{fig:sae_shift_commonclaim} track three layer-wise distances between the \textbf{truthful} centroid and its \textbf{deceptive} or \textbf{neutral} counterpart on LLaMA-3.1-8B-instruct. On both the templated \texttt{cities} set (Figure~\ref{fig:sae_shift_cities}) and the noisier \texttt{common\_claim\_true\_false} set (Figure~\ref{fig:sae_shift_commonclaim}), cosine similarity and feature-overlap plunge between layers 10–16. Meanwhile, the $\ell_2$ distance climbs to a clear peak. Deceptive prompts always induce a much larger shift than neutral prompts; the layer at which all three curves reach their extremum (Layer 16 for LLaMA, 21 for Gemma shown in Appendix\ref{appendix:sae_plots}) matches the peak in linear-probe accuracy. In contrast, the curves for \emph{truthful vs.\ neutral} stay almost flat, with $\text{cosine}>0.95$ and $\text{overlap}>0.80$ throughout.

\ding{175} \textbf{The geometric pattern of feature shifts is consistent regardless of dataset complexity, confirming a stereotyped truth–lie reorientation rather than dataset-specific noise.}
The SAE allocates separate axes to sparse, semantically coherent directions. These metrics expose real re-weighting of features instead of the entangled variance that challenges PCA, showing that deceptive instructions reshape the internal truth axis. \vspace{-4pt}

\subsubsection{Neuron-by-Neuron Analysis: Key Sparse Features Flip Sign}
While SAE feature shifts reveal robust geometric differences under different prompt types, they do not explain which specific neurons are responsible for these shifts. 
To localize which specific SAE directions drive the observed mid-layer shifts, for each layer, we identify the two sparse features whose activations differ most between \textbf{truthful} and \textbf{deceptive} inputs. Figure~\ref{fig:neuron_scatter_commonclaim} shows that, in \texttt{common\_claim\_true\_false}, these features exhibit a clear separation at layers 16 and 32: truthful and deceptive samples fall into distinct clusters along near-orthogonal directions, with minimal overlap. Similar trends are observed for another uncurated dataset \texttt{counterfact\_true\_false} (Appendix~\ref{appendix:sae_plots}). Violin plots (Figures~\ref{fig:violin_common_claim} and~\ref{fig:violin_counterfact}) confirm that the most responsive features show near-binary activation patterns, high for one instruction type and suppressed for the other. For example, top 2 features in Layer 16 is active almost exclusively under truthful prompts, while in Layer 32 this flips, activating strongly for deceptive inputs but not truthful ones.

\ding{176} \textbf{A small set of sparse features systematically flip their activation pattern between truthful and deceptive instructions.}  
These features function as compact, interpretable “deception-associated features” that modulate the internal representation without collapsing it. Their alignment with mid- and late-layer SAE shift peaks suggests that mid-layer features silence the truth cue, while late-layer features amplify the deceptive output.

\section{Related Work}
\label{sec:related_work}
Early studies showed that truth-related signals are encoded in activations and can be decoded via probes \citep{azaria2023internal, liu2024universaltruthfulnesshyperplaneinside, jin2025exploringconceptdepthlarge}. Further work uncovered linear structures underlying these representations \citep{marks2023geometry, ichmoukhamedov2025exploringgeneralizationllmtruth}, consistent with the Linear Representation Hypothesis \citep{park2023linear}. Various probing techniques, from Logistic Regression (LR) \citep{li2024inferencetimeinterventionelicitingtruthful, marks2023geometry} to polarity-aware approaches like TTPD \citep{burger2024truth}, have been used to find these “truth directions”, although generalization remains a challenge \citep{marks2023geometry, burger2024truth}. Some studies suggest that truth might be represented in a low-dimensional subspace rather than a single direction \citep{burger2024truth}. Beyond binary notions of truth, recent work shows that categorical and hierarchical concepts form simple polytopes (simplices) whose sub-components lie in orthogonal subspaces \citep{park2025the}. Related work above focuses on measuring or inducing truth-related directions within fixed models; orthogonally, model-efficiency transformations can alter these representations: pruning can be designed to preserve truthfulness \citep{fu2025pruning}, quantization may degrade or reshape truth-related behavior \citep{fu2025quantized}, and KV-cache compression aims to retain sequence information with minimal bias \citep{li2025faedkv}. We study a complementary axis—\emph{natural} instruction-induced shifts—holding architecture fixed.

Instruction-following behavior has also been linked to internal states \citep{heo2025llmsknowinternallyfollow}, with Representation Engineering \citep{zou2025representationengineeringtopdownapproach} and related methods demonstrate showing causal control over outputs \citep{li2024inferencetimeinterventionelicitingtruthful, marks2023geometry}, including knowledge-based refusals \citep{ferrando2025iknowentityknowledge}. Prompt-based approaches further show that truthfulness-relevant structure can be guided by input phrasing \citep{zhang2025promptguidedinternalstateshallucination}. We extend this by analyzing the \textit{natural} representational changes induced by different instruction types (truthful, neutral, deceptive), rather than externally manipulating them.

Superposition presents challenges for interpretability, and SAEs help isolate sparse, interpretable features \citep{bricken2023towards, cunningham2023sparse, shi2025routesparseautoencoderinterpret, cunningham2023sparseautoencodershighlyinterpretable}. Recent SAE releases like Gemma Scope \citep{lieberum2024gemma} and Llama Scope \citep{he2024llama} enable analysis in larger models. SAEs have been used to identify features tied to knowledge or behavior \citep{ferrando2025iknowentityknowledge, lan2025sparseautoencodersrevealuniversal}. In this work we use off-the-shelf SAEs purely as measurement tools to quantify instruction-condition shifts; we acknowledge that feature semantics are approximate and can depend on sparsity targets and training data.

\section{Conclusion}
This paper explored how deceptive instructions alter the internal representational geometry of LLMs compared to truthful or neutral ones. We found that the model's instructed ``True'' or ``Flase'' output is linearly decodable from intermediate activations across instruction types and datasets. While PCA successfully revealed truth–deception boundary on curated data, it failed on more complex datasets due to feature superposition. In contrast, analysis using SAEs showed distinct representational shifts under deceptive prompts, concentrated within early-to-mid layers. A neuron-level analysis further identified a few sparse features with polarity flips, serving as interpretable ``deception-associated features''. These insights clarify the internal geometry of instructed dishonesty in LLMs and offer a solid basis for future deception detection and model editing methods.

\section*{Limitations}
Our study is confined to English declaratives, frozen model weights, and linear probes. It neither tests causal interventions (e.g.\ activation patching) nor adversarial prompt recombinations.  
Furthermore, the evaluation data are labelled for binary factuality only; future work should extend to graded truth scales and multilingual settings.

\section*{Ethical Consideration}
Our research highlights the susceptibility of LLMs to produce falsehoods when exposed to carefully crafted prompts. This vulnerability raises concerns that a malicious user could exploit such behavior to propagate harmful or deceptive content. Nevertheless, we believe that current AI service providers prioritize truthfulness as a core objective in their deployment practices. Moreover, our deceptive prompts are intentionally constructed and easily identifiable, as they explicitly instruct LLMs to lie.

% \section*{Acknowledgments}

% This document has been adapted
% by Steven Bethard, Ryan Cotterell and Rui Yan
% from the instructions for earlier ACL and NAACL proceedings, including those for
% ACL 2019 by Douwe Kiela and Ivan Vuli\'{c},
% NAACL 2019 by Stephanie Lukin and Alla Roskovskaya,
% ACL 2018 by Shay Cohen, Kevin Gimpel, and Wei Lu,
% NAACL 2018 by Margaret Mitchell and Stephanie Lukin,
% Bib\TeX{} suggestions for (NA)ACL 2017/2018 from Jason Eisner,
% ACL 2017 by Dan Gildea and Min-Yen Kan,
% NAACL 2017 by Margaret Mitchell,
% ACL 2012 by Maggie Li and Michael White,
% ACL 2010 by Jing-Shin Chang and Philipp Koehn,
% ACL 2008 by Johanna D. Moore, Simone Teufel, James Allan, and Sadaoki Furui,
% ACL 2005 by Hwee Tou Ng and Kemal Oflazer,
% ACL 2002 by Eugene Charniak and Dekang Lin,
% and earlier ACL and EACL formats written by several people, including
% John Chen, Henry S. Thompson and Donald Walker.
% Additional elements were taken from the formatting instructions of the \emph{International Joint Conference on Artificial Intelligence} and the \emph{Conference on Computer Vision and Pattern Recognition}.

% \clearpage
\bibliography{main}

\begin{thebibliography}{53}
\providecommand{\natexlab}[1]{#1}

\bibitem[{Alain and Bengio(2018)}]{alain2018understandingintermediatelayersusing}
Guillaume Alain and Yoshua Bengio. 2018.
\newblock \href {https://arxiv.org/abs/1610.01644} {Understanding intermediate layers using linear classifier probes}.
\newblock \emph{Preprint}, arXiv:1610.01644.

\bibitem[{Azaria and Mitchell(2023{\natexlab{a}})}]{azaria2023internal}
Amos Azaria and Tom Mitchell. 2023{\natexlab{a}}.
\newblock The internal state of an {LLM} knows when it's lying.
\newblock In \emph{Findings of the Association for Computational Linguistics: EMNLP 2023}, pages 967--976, Singapore.

\bibitem[{Azaria and Mitchell(2023{\natexlab{b}})}]{azaria2023internalstatellmknows}
Amos Azaria and Tom Mitchell. 2023{\natexlab{b}}.
\newblock \href {https://arxiv.org/abs/2304.13734} {The internal state of an llm knows when it's lying}.
\newblock \emph{Preprint}, arXiv:2304.13734.

\bibitem[{Bricken et~al.(2023{\natexlab{a}})Bricken, Davies, Singh, Krotov, and Kreiman}]{bricken2023sparse}
Trenton Bricken, Xander Davies, Deepak Singh, Dmitry Krotov, and Gabriel Kreiman. 2023{\natexlab{a}}.
\newblock Sparse distributed memory is a continual learner.
\newblock \emph{arXiv preprint arXiv:2303.11934}.

\bibitem[{Bricken et~al.(2023{\natexlab{b}})Bricken, Templeton, Batson, Chen, Jermyn, Conerly, Turner, Anil, Denison, Askell et~al.}]{bricken2023towards}
Trenton Bricken, Adly Templeton, Joshua Batson, Brian Chen, Adam Jermyn, Tom Conerly, Nick Turner, Cem Anil, Carson Denison, Amanda Askell, and 1 others. 2023{\natexlab{b}}.
\newblock \href {https://transformer-circuits.pub/2023/monosemantic-features/index.html} {Towards monosemanticity: Decomposing language models with dictionary learning}.
\newblock Transformer Circuits Thread.

\bibitem[{Brown et~al.(2020)Brown, Mann, Ryder, Subbiah, Kaplan, Dhariwal, Neelakantan, Shyam, Sastry, Askell et~al.}]{brown2020language}
Tom Brown, Benjamin Mann, Nick Ryder, Melanie Subbiah, Jared~D Kaplan, Prafulla Dhariwal, Arvind Neelakantan, Pranav Shyam, Girish Sastry, Amanda Askell, and 1 others. 2020.
\newblock Language models are few-shot learners.
\newblock In \emph{Advances in neural information processing systems}, volume~33, pages 1877--1901.

\bibitem[{B{\"u}rger et~al.(2024)B{\"u}rger, Hamprecht, and Nadler}]{burger2024truth}
Lennart B{\"u}rger, Fred~A Hamprecht, and Boaz Nadler. 2024.
\newblock Truth is universal: Robust detection of lies in {LLMs}.
\newblock In \emph{Advances in Neural Information Processing Systems (NeurIPS)}.
\newblock ArXiv preprint arXiv:2407.12831v2.

\bibitem[{Casper et~al.(2023)Casper, Lin, Kwon, Culp, and Hadfield-Menell}]{casper2023explore}
Stephen Casper, Jason Lin, Joe Kwon, Gatlen Culp, and Dylan Hadfield-Menell. 2023.
\newblock \href {https://arxiv.org/abs/2306.09442} {Explore, establish, exploit: Red teaming language models from scratch}.
\newblock \emph{Preprint}, arXiv:2306.09442.

\bibitem[{Chojnacki(2025)}]{chojnacki2025interpretableriskmitigationllm}
Jan Chojnacki. 2025.
\newblock \href {https://arxiv.org/abs/2505.10670} {Interpretable risk mitigation in llm agent systems}.
\newblock \emph{Preprint}, arXiv:2505.10670.

\bibitem[{Cunningham et~al.(2023{\natexlab{a}})Cunningham, Ewart, Riggs, Huben, and Sharkey}]{cunningham2023sparse}
Hoagy Cunningham, Aidan Ewart, Logan Riggs, Robert Huben, and Lee Sharkey. 2023{\natexlab{a}}.
\newblock \href {https://arxiv.org/abs/2309.08600} {Sparse autoencoders find highly interpretable features in language models}.
\newblock \emph{Preprint}, arXiv:2309.08600.

\bibitem[{Cunningham et~al.(2023{\natexlab{b}})Cunningham, Ewart, Riggs, Huben, and Sharkey}]{cunningham2023sparseautoencodershighlyinterpretable}
Hoagy Cunningham, Aidan Ewart, Logan Riggs, Robert Huben, and Lee Sharkey. 2023{\natexlab{b}}.
\newblock \href {https://arxiv.org/abs/2309.08600} {Sparse autoencoders find highly interpretable features in language models}.
\newblock \emph{Preprint}, arXiv:2309.08600.

\bibitem[{Dinan et~al.(2019)Dinan, Roller, Shuster, Fan, Auli, and Weston}]{dinan2019wizardwikipediaknowledgepoweredconversational}
Emily Dinan, Stephen Roller, Kurt Shuster, Angela Fan, Michael Auli, and Jason Weston. 2019.
\newblock \href {https://arxiv.org/abs/1811.01241} {Wizard of wikipedia: Knowledge-powered conversational agents}.
\newblock \emph{Preprint}, arXiv:1811.01241.

\bibitem[{Dreyer et~al.(2024)Dreyer, Purelku, Vielhaben, Samek, and Lapuschkin}]{dreyer2024pureturningpolysemanticneurons}
Maximilian Dreyer, Erblina Purelku, Johanna Vielhaben, Wojciech Samek, and Sebastian Lapuschkin. 2024.
\newblock \href {https://arxiv.org/abs/2404.06453} {Pure: Turning polysemantic neurons into pure features by identifying relevant circuits}.
\newblock \emph{Preprint}, arXiv:2404.06453.

\bibitem[{Elhage et~al.(2022)Elhage, Hume, Olsson, Schiefer, Henighan, Kravec, Hatfield-Dodds, Lasenby, Drain, Chen, Grosse, McCandlish, Kaplan, Amodei, Wattenberg, and Olah}]{elhage2022toymodelssuperposition}
Nelson Elhage, Tristan Hume, Catherine Olsson, Nicholas Schiefer, Tom Henighan, Shauna Kravec, Zac Hatfield-Dodds, Robert Lasenby, Dawn Drain, Carol Chen, Roger Grosse, Sam McCandlish, Jared Kaplan, Dario Amodei, Martin Wattenberg, and Christopher Olah. 2022.
\newblock \href {https://arxiv.org/abs/2209.10652} {Toy models of superposition}.
\newblock \emph{Preprint}, arXiv:2209.10652.

\bibitem[{Ferrando et~al.(2025)Ferrando, Obeso, Rajamanoharan, and Nanda}]{ferrando2025iknowentityknowledge}
Javier Ferrando, Oscar Obeso, Senthooran Rajamanoharan, and Neel Nanda. 2025.
\newblock \href {https://arxiv.org/abs/2411.14257} {Do i know this entity? knowledge awareness and hallucinations in language models}.
\newblock \emph{Preprint}, arXiv:2411.14257.

\bibitem[{Fu et~al.(2025{\natexlab{a}})Fu, Li, Long, Yu, Han, Yin, and Li}]{fu2025pruning}
Yao Fu, Runchao Li, Xianxuan Long, Haotian Yu, Xiaotian Han, Yu~Yin, and Pan Li. 2025{\natexlab{a}}.
\newblock Pruning weights but not truth: Safeguarding truthfulness while pruning llms.
\newblock \emph{arXiv preprint arXiv:2509.00096}.

\bibitem[{Fu et~al.(2025{\natexlab{b}})Fu, Long, Li, Yu, Sheng, Han, Yin, and Li}]{fu2025quantized}
Yao Fu, Xianxuan Long, Runchao Li, Haotian Yu, Mu~Sheng, Xiaotian Han, Yu~Yin, and Pan Li. 2025{\natexlab{b}}.
\newblock Quantized but deceptive? a multi-dimensional truthfulness evaluation of quantized llms.
\newblock \emph{arXiv preprint arXiv:2508.19432}.

\bibitem[{Greenblatt et~al.(2024)Greenblatt, Denison, Wright, Roger, MacDiarmid, Marks, Treutlein, Belonax, Chen, Duvenaud, Khan, Michael, Mindermann, Perez, Petrini, Uesato, Kaplan, Shlegeris, Bowman, and Hubinger}]{greenblatt2024alignmentfakinglargelanguage}
Ryan Greenblatt, Carson Denison, Benjamin Wright, Fabien Roger, Monte MacDiarmid, Sam Marks, Johannes Treutlein, Tim Belonax, Jack Chen, David Duvenaud, Akbir Khan, Julian Michael, Sören Mindermann, Ethan Perez, Linda Petrini, Jonathan Uesato, Jared Kaplan, Buck Shlegeris, Samuel~R. Bowman, and Evan Hubinger. 2024.
\newblock \href {https://arxiv.org/abs/2412.14093} {Alignment faking in large language models}.
\newblock \emph{Preprint}, arXiv:2412.14093.

\bibitem[{Hagendorff(2024)}]{Hagendorff_2024}
Thilo Hagendorff. 2024.
\newblock \href {https://doi.org/10.1073/pnas.2317967121} {Deception abilities emerged in large language models}.
\newblock \emph{Proceedings of the National Academy of Sciences}, 121(24).

\bibitem[{He et~al.(2024)He, Shu, Ge, Chen, Wang, Zhou, Guo, Huang, Wu, Liu et~al.}]{he2024llama}
Zhengfu He, Wentao Shu, Xuyang Ge, Lingjie Chen, Junxuan Wang, Yunhua Zhou, Qipeng Guo, Xuanjing Huang, Zuxuan Wu, Frances Liu, and 1 others. 2024.
\newblock \href {https://arxiv.org/abs/2410.20526} {Llama scope: Extracting millions of features from llama-3.1-8b with sparse autoencoders}.
\newblock \emph{Preprint}, arXiv:2410.20526.

\bibitem[{Heo et~al.(2025)Heo, Heinze-Deml, Elachqar, Chan, Ren, Nallasamy, Miller, and Narain}]{heo2025llmsknowinternallyfollow}
Juyeon Heo, Christina Heinze-Deml, Oussama Elachqar, Kwan Ho~Ryan Chan, Shirley Ren, Udhay Nallasamy, Andy Miller, and Jaya Narain. 2025.
\newblock \href {https://arxiv.org/abs/2410.14516} {Do llms "know" internally when they follow instructions?}
\newblock \emph{Preprint}, arXiv:2410.14516.

\bibitem[{Ichmoukhamedov and Martens(2025)}]{ichmoukhamedov2025exploringgeneralizationllmtruth}
Timour Ichmoukhamedov and David Martens. 2025.
\newblock \href {https://arxiv.org/abs/2505.09807} {Exploring the generalization of llm truth directions on conversational formats}.
\newblock \emph{Preprint}, arXiv:2505.09807.

\bibitem[{Jiang et~al.(2023)Jiang, Sablayrolles, Mensch, Bamford, Chaplot, de~las Casas, Bressand, Lengyel, Lample, Saulnier, Lavaud, Lachaux, Stock, Scao, Lavril, Wang, Lacroix, and Sayed}]{jiang2023mistral7b}
Albert~Q. Jiang, Alexandre Sablayrolles, Arthur Mensch, Chris Bamford, Devendra~Singh Chaplot, Diego de~las Casas, Florian Bressand, Gianna Lengyel, Guillaume Lample, Lucile Saulnier, Lélio~Renard Lavaud, Marie-Anne Lachaux, Pierre Stock, Teven~Le Scao, Thibaut Lavril, Thomas Wang, Timothée Lacroix, and William~El Sayed. 2023.
\newblock \href {https://arxiv.org/abs/2310.06825} {Mistral 7b}.
\newblock \emph{Preprint}, arXiv:2310.06825.

\bibitem[{Jin et~al.(2025)Jin, Yu, Huang, Zeng, Wang, Hua, Zhao, Mei, Meng, Ding, Yang, Du, and Zhang}]{jin2025exploringconceptdepthlarge}
Mingyu Jin, Qinkai Yu, Jingyuan Huang, Qingcheng Zeng, Zhenting Wang, Wenyue Hua, Haiyan Zhao, Kai Mei, Yanda Meng, Kaize Ding, Fan Yang, Mengnan Du, and Yongfeng Zhang. 2025.
\newblock \href {https://arxiv.org/abs/2404.07066} {Exploring concept depth: How large language models acquire knowledge and concept at different layers?}
\newblock \emph{Preprint}, arXiv:2404.07066.

\bibitem[{Khatun and Brown(2024)}]{khatun2024truthevaldatasetevaluatellm}
Aisha Khatun and Daniel~G. Brown. 2024.
\newblock \href {https://arxiv.org/abs/2406.01855} {Trutheval: A dataset to evaluate llm truthfulness and reliability}.
\newblock \emph{Preprint}, arXiv:2406.01855.

\bibitem[{Lan et~al.(2025)Lan, Torr, Meek, Khakzar, Krueger, and Barez}]{lan2025sparseautoencodersrevealuniversal}
Michael Lan, Philip Torr, Austin Meek, Ashkan Khakzar, David Krueger, and Fazl Barez. 2025.
\newblock \href {https://arxiv.org/abs/2410.06981} {Sparse autoencoders reveal universal feature spaces across large language models}.
\newblock \emph{Preprint}, arXiv:2410.06981.

\bibitem[{Li et~al.(2024)Li, Patel, Viégas, Pfister, and Wattenberg}]{li2024inferencetimeinterventionelicitingtruthful}
Kenneth Li, Oam Patel, Fernanda Viégas, Hanspeter Pfister, and Martin Wattenberg. 2024.
\newblock \href {https://arxiv.org/abs/2306.03341} {Inference-time intervention: Eliciting truthful answers from a language model}.
\newblock \emph{Preprint}, arXiv:2306.03341.

\bibitem[{Li et~al.(2025)Li, Fu, Sheng, Long, Yu, and Li}]{li2025faedkv}
Runchao Li, Yao Fu, Mu~Sheng, Xianxuan Long, Haotian Yu, and Pan Li. 2025.
\newblock Faedkv: Infinite-window fourier transform for unbiased kv cache compression.
\newblock \emph{arXiv preprint arXiv:2507.20030}.

\bibitem[{Lieberum et~al.(2024)Lieberum, Rajamanoharan, Conmy, Smith, Sonnerat, Varma, Kramár, Dragan, Shah, and Nanda}]{lieberum2024gemma}
Tom Lieberum, Senthooran Rajamanoharan, Arthur Conmy, Lewis Smith, Nicolas Sonnerat, Vikrant Varma, János Kramár, Anca Dragan, Rohin Shah, and Neel Nanda. 2024.
\newblock \href {https://arxiv.org/abs/2408.05147} {Gemma scope: Open sparse autoencoders everywhere all at once on gemma 2}.
\newblock \emph{Preprint}, arXiv:2408.05147.

\bibitem[{Lin(2023)}]{neuronpedia}
Johnny Lin. 2023.
\newblock \href {https://www.neuronpedia.org} {Neuronpedia: Interactive reference and tooling for analyzing neural networks}.
\newblock Software available from neuronpedia.org.

\bibitem[{Lin et~al.(2022)Lin, Hilton, and Evans}]{lin2022truthfulqameasuringmodelsmimic}
Stephanie Lin, Jacob Hilton, and Owain Evans. 2022.
\newblock \href {https://arxiv.org/abs/2109.07958} {Truthfulqa: Measuring how models mimic human falsehoods}.
\newblock \emph{Preprint}, arXiv:2109.07958.

\bibitem[{Liu et~al.(2024)Liu, Chen, Cheng, and He}]{liu2024universaltruthfulnesshyperplaneinside}
Junteng Liu, Shiqi Chen, Yu~Cheng, and Junxian He. 2024.
\newblock \href {https://arxiv.org/abs/2407.08582} {On the universal truthfulness hyperplane inside llms}.
\newblock \emph{Preprint}, arXiv:2407.08582.

\bibitem[{Marks and Tegmark(2024)}]{marks2023geometry}
Samuel Marks and Max Tegmark. 2024.
\newblock The geometry of truth: Emergent linear structure in {LLM} representations of true/false datasets.
\newblock In \emph{Conference on Language Modeling (COLM)}.
\newblock ArXiv preprint arXiv:2310.06824v3, Published as a conference paper at COLM 2024.

\bibitem[{Meng et~al.(2023)Meng, Bau, Andonian, and Belinkov}]{meng2023locatingeditingfactualassociations}
Kevin Meng, David Bau, Alex Andonian, and Yonatan Belinkov. 2023.
\newblock \href {https://arxiv.org/abs/2202.05262} {Locating and editing factual associations in gpt}.
\newblock \emph{Preprint}, arXiv:2202.05262.

\bibitem[{Pacchiardi et~al.(2023)Pacchiardi, Chan, Mindermann, Moscovitz, Pan, Gal, Evans, and Brauner}]{pacchiardi2023catchailiarlie}
Lorenzo Pacchiardi, Alex~J. Chan, Sören Mindermann, Ilan Moscovitz, Alexa~Y. Pan, Yarin Gal, Owain Evans, and Jan Brauner. 2023.
\newblock \href {https://arxiv.org/abs/2309.15840} {How to catch an ai liar: Lie detection in black-box llms by asking unrelated questions}.
\newblock \emph{Preprint}, arXiv:2309.15840.

\bibitem[{Park et~al.(2025)Park, Choe, Jiang, and Veitch}]{park2025the}
Kiho Park, Yo~Joong Choe, Yibo Jiang, and Victor Veitch. 2025.
\newblock \href {https://openreview.net/forum?id=bVTM2QKYuA} {The geometry of categorical and hierarchical concepts in large language models}.
\newblock In \emph{The Thirteenth International Conference on Learning Representations}.

\bibitem[{Park et~al.(2024)Park, Choe, and Veitch}]{park2023linear}
Kiho Park, Yo~Joong Choe, and Victor Veitch. 2024.
\newblock The linear representation hypothesis and the geometry of large language models.
\newblock In \emph{Proceedings of the 41st International Conference on Machine Learning (ICML)}, volume 235 of \emph{PMLR}.
\newblock ArXiv preprint arXiv:2311.03658v2.

\bibitem[{Qin et~al.(2024)Qin, Song, Hu, Yao, Cho, Wang, Wu, Liu, Liu, and Yu}]{qin2024infobenchevaluatinginstructionfollowing}
Yiwei Qin, Kaiqiang Song, Yebowen Hu, Wenlin Yao, Sangwoo Cho, Xiaoyang Wang, Xuansheng Wu, Fei Liu, Pengfei Liu, and Dong Yu. 2024.
\newblock \href {https://arxiv.org/abs/2401.03601} {Infobench: Evaluating instruction following ability in large language models}.
\newblock \emph{Preprint}, arXiv:2401.03601.

\bibitem[{Qwen et~al.(2025)Qwen, :, Yang, Yang, Zhang, Hui, Zheng, Yu, Li, Liu, Huang, Wei, Lin, Yang, Tu, Zhang, Yang, Yang, Zhou, Lin, Dang, Lu, Bao, Yang, Yu, Li, Xue, Zhang, Zhu, Men, Lin, Li, Tang, Xia, Ren, Ren, Fan, Su, Zhang, Wan, Liu, Cui, Zhang, and Qiu}]{qwen2025qwen25technicalreport}
Qwen, :, An~Yang, Baosong Yang, Beichen Zhang, Binyuan Hui, Bo~Zheng, Bowen Yu, Chengyuan Li, Dayiheng Liu, Fei Huang, Haoran Wei, Huan Lin, Jian Yang, Jianhong Tu, Jianwei Zhang, Jianxin Yang, Jiaxi Yang, Jingren Zhou, and 25 others. 2025.
\newblock \href {https://arxiv.org/abs/2412.15115} {Qwen2.5 technical report}.
\newblock \emph{Preprint}, arXiv:2412.15115.

\bibitem[{Scheurer et~al.(2024)Scheurer, Balesni, and Hobbhahn}]{scheurer2024large}
J{\'e}r{\'e}my Scheurer, Mikita Balesni, and Marius Hobbhahn. 2024.
\newblock \href {https://openreview.net/forum?id=HduMpot9sJ} {Large language models can strategically deceive their users when put under pressure}.
\newblock In \emph{ICLR 2024 Workshop on Large Language Model (LLM) Agents}.

\bibitem[{Shah et~al.(2025)Shah, Irpan, Turner, Wang, Conmy, Lindner, Brown-Cohen, Ho, Nanda, Popa, Jain, Greig, Albanie, Emmons, Farquhar, Krier, Rajamanoharan, Bridgers, Ijitoye, Everitt, Krakovna, Varma, Mikulik, Kenton, Orr, Legg, Goodman, Dafoe, Flynn, and Dragan}]{shah2025approachtechnicalagisafety}
Rohin Shah, Alex Irpan, Alexander~Matt Turner, Anna Wang, Arthur Conmy, David Lindner, Jonah Brown-Cohen, Lewis Ho, Neel Nanda, Raluca~Ada Popa, Rishub Jain, Rory Greig, Samuel Albanie, Scott Emmons, Sebastian Farquhar, Sébastien Krier, Senthooran Rajamanoharan, Sophie Bridgers, Tobi Ijitoye, and 11 others. 2025.
\newblock \href {https://arxiv.org/abs/2504.01849} {An approach to technical agi safety and security}.
\newblock \emph{Preprint}, arXiv:2504.01849.

\bibitem[{Sharkey et~al.(2025)Sharkey, Chughtai, Batson, Lindsey, Wu, Bushnaq, Goldowsky-Dill, Heimersheim, Ortega, Bloom, Biderman, Garriga-Alonso, Conmy, Nanda, Rumbelow, Wattenberg, Schoots, Miller, Michaud, Casper, Tegmark, Saunders, Bau, Todd, Geiger, Geva, Hoogland, Murfet, and McGrath}]{sharkey2025openproblemsmechanisticinterpretability}
Lee Sharkey, Bilal Chughtai, Joshua Batson, Jack Lindsey, Jeff Wu, Lucius Bushnaq, Nicholas Goldowsky-Dill, Stefan Heimersheim, Alejandro Ortega, Joseph Bloom, Stella Biderman, Adria Garriga-Alonso, Arthur Conmy, Neel Nanda, Jessica Rumbelow, Martin Wattenberg, Nandi Schoots, Joseph Miller, Eric~J. Michaud, and 10 others. 2025.
\newblock \href {https://arxiv.org/abs/2501.16496} {Open problems in mechanistic interpretability}.
\newblock \emph{Preprint}, arXiv:2501.16496.

\bibitem[{Shen and Younes(2024)}]{shen2024reimagininglinearprobingkolmogorovarnold}
Sheng Shen and Rabih Younes. 2024.
\newblock \href {https://arxiv.org/abs/2409.07763} {Reimagining linear probing: Kolmogorov-arnold networks in transfer learning}.
\newblock \emph{Preprint}, arXiv:2409.07763.

\bibitem[{Shi et~al.(2025)Shi, Li, Liang, Wan, Ma, Wang, and He}]{shi2025routesparseautoencoderinterpret}
Wei Shi, Sihang Li, Tao Liang, Mingyang Wan, Guojun Ma, Xiang Wang, and Xiangnan He. 2025.
\newblock \href {https://arxiv.org/abs/2503.08200} {Route sparse autoencoder to interpret large language models}.
\newblock \emph{Preprint}, arXiv:2503.08200.

\bibitem[{Shu et~al.(2025)Shu, Wu, Zhao, Rai, Yao, Liu, and Du}]{shu2025surveysparseautoencodersinterpreting}
Dong Shu, Xuansheng Wu, Haiyan Zhao, Daking Rai, Ziyu Yao, Ninghao Liu, and Mengnan Du. 2025.
\newblock \href {https://arxiv.org/abs/2503.05613} {A survey on sparse autoencoders: Interpreting the internal mechanisms of large language models}.
\newblock \emph{Preprint}, arXiv:2503.05613.

\bibitem[{Team et~al.(2024)Team, Riviere, Pathak, Sessa, Hardin, Bhupatiraju, Hussenot, Mesnard, Shahriari, Ramé, Ferret, Liu, Tafti, Friesen, Casbon, Ramos, Kumar, Lan, Jerome, Tsitsulin, Vieillard, Stanczyk, Girgin, Momchev, Hoffman, Thakoor, Grill, Neyshabur, Bachem, Walton, Severyn, Parrish, Ahmad, Hutchison, Abdagic, Carl, Shen, Brock, Coenen, Laforge, Paterson, Bastian, Piot, Wu, Royal, Chen, Kumar, Perry, Welty, Choquette-Choo, Sinopalnikov, Weinberger, Vijaykumar, Rogozińska, Herbison, Bandy, Wang, Noland, Moreira, Senter, Eltyshev, Visin, Rasskin, Wei, Cameron, Martins, Hashemi, Klimczak-Plucińska, Batra, Dhand, Nardini, Mein, Zhou, Svensson, Stanway, Chan, Zhou, Carrasqueira, Iljazi, Becker, Fernandez, van Amersfoort, Gordon, Lipschultz, Newlan, yeong Ji, Mohamed, Badola, Black, Millican, McDonell, Nguyen, Sodhia, Greene, Sjoesund, Usui, Sifre, Heuermann, Lago, McNealus, Soares, Kilpatrick, Dixon, Martins, Reid, Singh, Iverson, Görner, Velloso, Wirth, Davidow, Miller, Rahtz, Watson, Risdal,
  Kazemi, Moynihan, Zhang, Kahng, Park, Rahman, Khatwani, Dao, Bardoliwalla, Devanathan, Dumai, Chauhan, Wahltinez, Botarda, Barnes, Barham, Michel, Jin, Georgiev, Culliton, Kuppala, Comanescu, Merhej, Jana, Rokni, Agarwal, Mullins, Saadat, Carthy, Cogan, Perrin, Arnold, Krause, Dai, Garg, Sheth, Ronstrom, Chan, Jordan, Yu, Eccles, Hennigan, Kocisky, Doshi, Jain, Yadav, Meshram, Dharmadhikari, Barkley, Wei, Ye, Han, Kwon, Xu, Shen, Gong, Wei, Cotruta, Kirk, Rao, Giang, Peran, Warkentin, Collins, Barral, Ghahramani, Hadsell, Sculley, Banks, Dragan, Petrov, Vinyals, Dean, Hassabis, Kavukcuoglu, Farabet, Buchatskaya, Borgeaud, Fiedel, Joulin, Kenealy, Dadashi, and Andreev}]{gemmateam2024gemma2improvingopen}
Gemma Team, Morgane Riviere, Shreya Pathak, Pier~Giuseppe Sessa, Cassidy Hardin, Surya Bhupatiraju, Léonard Hussenot, Thomas Mesnard, Bobak Shahriari, Alexandre Ramé, Johan Ferret, Peter Liu, Pouya Tafti, Abe Friesen, Michelle Casbon, Sabela Ramos, Ravin Kumar, Charline~Le Lan, Sammy Jerome, and 179 others. 2024.
\newblock \href {https://arxiv.org/abs/2408.00118} {Gemma 2: Improving open language models at a practical size}.
\newblock \emph{Preprint}, arXiv:2408.00118.

\bibitem[{Tomihari and Sato(2024)}]{tomihari2024understandinglinearprobingfinetuning}
Akiyoshi Tomihari and Issei Sato. 2024.
\newblock \href {https://arxiv.org/abs/2405.16747} {Understanding linear probing then fine-tuning language models from ntk perspective}.
\newblock \emph{Preprint}, arXiv:2405.16747.

\bibitem[{Touvron et~al.(2023)Touvron, Lavril, Izacard, Martinet, Lachaux, Lacroix, Rozi{\`e}re, Goyal, Hambro, Azhar et~al.}]{touvron2023llama}
Hugo Touvron, Thibaut Lavril, Gautier Izacard, Xavier Martinet, Marie-Anne Lachaux, Timoth{\'e}e Lacroix, Baptiste Rozi{\`e}re, Naman Goyal, Eric Hambro, Faisal Azhar, and 1 others. 2023.
\newblock \href {https://arxiv.org/abs/2302.13971} {Llama: Open and efficient foundation language models}.
\newblock \emph{Preprint}, arXiv:2302.13971.

\bibitem[{Wu et~al.(2025)Wu, Pan, Hong, and Yang}]{wu2025opendeceptionbenchmarkinginvestigatingai}
Yichen Wu, Xudong Pan, Geng Hong, and Min Yang. 2025.
\newblock \href {https://arxiv.org/abs/2504.13707} {Opendeception: Benchmarking and investigating ai deceptive behaviors via open-ended interaction simulation}.
\newblock \emph{Preprint}, arXiv:2504.13707.

\bibitem[{Zhang et~al.(2025)Zhang, Yu, Yi, Zhang, Li, and Liu}]{zhang2025promptguidedinternalstateshallucination}
Fujie Zhang, Peiqi Yu, Biao Yi, Baolei Zhang, Tong Li, and Zheli Liu. 2025.
\newblock \href {https://arxiv.org/abs/2411.04847} {Prompt-guided internal states for hallucination detection of large language models}.
\newblock \emph{Preprint}, arXiv:2411.04847.

\bibitem[{Zhang et~al.(2022)Zhang, Roller, Goyal, Artetxe, Chen, Chen, Dewan, Diab, Li, Lin, Mihaylov, Ott, Shleifer, Shuster, Simig, Koura, Sridhar, Wang, and Zettlemoyer}]{zhang2022optopenpretrainedtransformer}
Susan Zhang, Stephen Roller, Naman Goyal, Mikel Artetxe, Moya Chen, Shuohui Chen, Christopher Dewan, Mona Diab, Xian Li, Xi~Victoria Lin, Todor Mihaylov, Myle Ott, Sam Shleifer, Kurt Shuster, Daniel Simig, Punit~Singh Koura, Anjali Sridhar, Tianlu Wang, and Luke Zettlemoyer. 2022.
\newblock \href {https://arxiv.org/abs/2205.01068} {Opt: Open pre-trained transformer language models}.
\newblock \emph{Preprint}, arXiv:2205.01068.

\bibitem[{Zhou et~al.(2023)Zhou, Lu, Mishra, Brahma, Basu, Luan, Zhou, and Hou}]{zhou2023instructionfollowingevaluationlargelanguage}
Jeffrey Zhou, Tianjian Lu, Swaroop Mishra, Siddhartha Brahma, Sujoy Basu, Yi~Luan, Denny Zhou, and Le~Hou. 2023.
\newblock \href {https://arxiv.org/abs/2311.07911} {Instruction-following evaluation for large language models}.
\newblock \emph{Preprint}, arXiv:2311.07911.

\bibitem[{Zou et~al.(2025)Zou, Phan, Chen, Campbell, Guo, Ren, Pan, Yin, Mazeika, Dombrowski, Goel, Li, Byun, Wang, Mallen, Basart, Koyejo, Song, Fredrikson, Kolter, and Hendrycks}]{zou2025representationengineeringtopdownapproach}
Andy Zou, Long Phan, Sarah Chen, James Campbell, Phillip Guo, Richard Ren, Alexander Pan, Xuwang Yin, Mantas Mazeika, Ann-Kathrin Dombrowski, Shashwat Goel, Nathaniel Li, Michael~J. Byun, Zifan Wang, Alex Mallen, Steven Basart, Sanmi Koyejo, Dawn Song, Matt Fredrikson, and 2 others. 2025.
\newblock \href {https://arxiv.org/abs/2310.01405} {Representation engineering: A top-down approach to ai transparency}.
\newblock \emph{Preprint}, arXiv:2310.01405.

\end{thebibliography}

\clearpage
\appendix
\section{Dataset Details}
\label{app:data}
This appendix documents every corpus used in our experiments, including
its provenance, construction protocol, and basic statistics.\footnote{All
CSV files, generation scripts and pre-processed activation matrices will
be released upon publication.}  We partition the resources into
\emph{Curated Logical-Bench} (§\ref{app:curated}) and
\emph{Open-Domain Fact-Bench} (§\ref{app:open}).  The former is further
broken down into (i) \emph{topic-specific} domains with four logical
variants and (ii) two \emph{relational comparison} sets.

\subsection{Curated Logical-Bench}
\label{app:curated}

\begin{table*}[h]
    \centering
    \caption{Our datasets \(D_i\)}
    \label{tab:datasets}
    \begin{tabular}{lll}
        \toprule
        \textbf{Name} & \textbf{Description} & \textbf{Rows} \\
        \midrule
        \texttt{cities}         & "The city of [\texttt{city}] is in [\texttt{country}]."  & 1496 \\
        \texttt{sp\_en\_trans}  & "The Spanish word ‘[\texttt{word}]' means ‘[\texttt{English word}]'."  & 354 \\
        \texttt{element\_symb}  & "[\texttt{element}] has the symbol of [\texttt{symbol}]."   & 186 \\
        \texttt{animal\_class}  &  "The [\texttt{animal}] is a [\texttt{animal\_class}]." & 164 \\
        \texttt{inventors}      & "[\texttt{inventor}] lived in [\texttt{counrty}]." & 406 \\
        \texttt{facts}          & Diverse scientific facts & 561 \\
        \midrule
        \texttt{larger\_than}   &  "$\textit{x}$ is larger than $\textit{y}$."  & 1980 \\
        \texttt{smaller\_than}   &  "$\textit{x}$ is smaller than $\textit{y}$." & 1980 \\
        \midrule
        \texttt{common\_claim\_true\_false}   & Various claims; from \cite{azaria2023internal} & 4450\\
        \texttt{counterfact\_true\_false}    & Various factual recall claims; from \cite{meng2023locatingeditingfactualassociations} & 31960\\

        \bottomrule
    \end{tabular}
\end{table*}

\paragraph{Affirmative Statements}  \hspace{5pt} \citet{burger2024truth} collect six topic specific datasets of affirmative statements, each on a single topic as detailed in Table \ref{tab:datasets}. The \texttt{cities} and \texttt{sp\_en\_trans} datasets are from \citet{marks2023geometry}, while \texttt{element\_symb}, \texttt{animal\_class}, \texttt{inventors} and \texttt{facts} are subsets of the datasets compiled by \citet{azaria2023internal}. All datasets, with the exception of "facts", consist of simple, uncontroversial and unambiguous statements. Each dataset (except "facts") follows a consistent template. For example, the template of \texttt{cities} is "The city of <city name> is in <country name>.", whereas that of \texttt{sp\_en\_trans} is "The Spanish word <Spanish word> means <English word>." In contrast, "facts" is more diverse, containing statements of various forms and topics. 

\paragraph{Negated Statements.} \hspace{5pt} Following \citet{burger2024truth}, in this paper, each of the statements in the six datasets from Table~\ref{tab:datasets} is negated by inserting the word "not". For instance, "The Spanish word 'dos' means 'enemy'." (False) turns into "The Spanish word 'dos' does not mean 'enemy'." (True). This results in six additional datasets of negated statements, denoted by the prefix "neg\_".

\paragraph{Logical Conjunctions.} \hspace{5pt} We use the following template to generate the logical conjunctions from six datasets in Table \ref{tab:datasets}, separately for each topic:
\begin{itemize}
    \item It is the case both that [statement 1] and that [statement 2].
\end{itemize}

Following the recent work \cite{burger2024truth}, the two statements are sampled independently to be true with probability $\frac{1}{\sqrt{2}}$. This ensures that the overall dataset is balanced between true and false statements, but that there is no statistical dependency between the truth of the first and second statement in the conjunction. The new datasets are denoted by the suffix "\_conj", e.g., \texttt{sp\_en\_trans\_conj} or \texttt{facts\_conj}. Each dataset contains 500 statements. Examples include:
\begin{itemize}
    \item It is the case both that the city of Al Ain City is in the United Arab Emirates and that the city of Jilin is in China. (True)
    \item It is the case both that Oxygen is necessary for humans to breathe and that the sun revolves around the moon. (False)
\end{itemize}

\paragraph{Logical Disjunctions.}\hspace{5pt}
The templates for the disjunctions were adapted to each dataset in Table \ref{tab:datasets}, combining two statements as follows:
\begin{itemize}
    \item \texttt{cities\_disj}: It is the case either that the city of [city 1] is in [country 1/2] or that it is in [country 2/1].
    \item \texttt{sp\_en\_trans\_disj}: It is the case either that the Spanish word [Spanish word 1] means [English word 1/2] or that it means [English word 2/1].
\end{itemize}

Analogous templates were all used for rest of datasets \texttt{element\_symb}, \texttt{inventors}, and \texttt{animal\_class}. \citet{burger2024truth} sample the first statement to be true with a probability of $1/2$ and then sample a second statement, ensuring the end-word (e.g., [country 2]) would be incorrect for statement 1. The order of the two end-words is flipped with a probability of $1/2$. The new datasets are denoted by the suffix "\_disj", e.g., \texttt{sp\_en\_trans\_disj}, and each contains 500 statements. Examples include:
\begin{itemize}
    \item It is the case either that the city of Korla is in Azerbaijan or that it is in Russia. (False)
    \item It is the case either that the Spanish word ‘carne’ means ‘meat’ or that it means ‘seven’. (True)
    \item It is the case either that Bromine has the symbol Ce or that it has the symbol Mo. (False)
\end{itemize}

Combining statements in this simple way is not possible for the more diverse \texttt{facts} dataset and \citet{burger2024truth} use the following template instead:
\begin{itemize}
    \item It is the case either that [statement 1] or that [statement 2].
\end{itemize}

Following \citet{burger2024truth}, we sample the two statements independently to be true with probability $1 - \frac{1}{\sqrt{2}}$. This ensures that the overall dataset is balanced between true and false statements, but that there is no statistical dependency between the truth of the first and second statement in the disjunction. Examples include:
\begin{itemize}
    \item It is the case either that the Earth is the third planet from the sun or that the Milky Way is a linear galaxy. (True)
    \item It is the case either that the fastest bird in the world is the penguin or that Oxygen is harmful to human breathing. (False)
\end{itemize}

\subsection{Open-Domain Fact-Bench}
\label{app:open}
\paragraph{\texttt{common\_claim\_true\_false}} \hspace{5pt} 
\texttt{CommonClaim} is introduced by \citet{casper2023explore}, containing 20,000 GPT-3-text-davinci-002 generations which are labeled as true, false, or neither, according to human common knowledge. \citet{marks2023geometry} adapted \texttt{CommonClaim} by selecting statements labeled true or false, then removing excess true statements to balance the dataset. This modified version consists of 4450 statements. Example statements:
\begin{itemize}
    \item Bananas are believed to be one of the oldest fruits in the world. (True)
    \item Crazy ants have taken over Cape Canaveral. (False)
\end{itemize}

\paragraph{\texttt{counterfact\_true\_false}} \hspace{5pt} 
\texttt{Counterfact} was introduced in \citet{meng2023locatingeditingfactualassociations} and consists of factual recall statements. We adapt Counterfact by using statements which form complete sentences and, for each such statement, using both the true version and a false version given by one of Counterfact’s suggested false modifications. We also append a period to the end. Example statements:
\begin{itemize}
    \item Olaus Rudbeck spoke the language Swedish. (True)
    \item The official religion of Malacca sultanate is Christianity. (False)
\end{itemize}

\section{Complete Layer-wise Probing Results}
\label{appendix:probe_table}
This appendix aims to provide a more exhaustive quantitative analysis of the internal representations within the LLMs under investigation. Specifically, Table~\ref{tab:layer-wise_prob_llama} and Table~\ref{tab:layer-wise_prob_gemma} present the complete accuracy information from the layer-wise probing conducted on LLaMA-3.1-8B-IT and Gemma2-9B-IT. These results span the three distinct instructional conditions—Truthful, Neutral, and Deceptive prompts—and utilize two different probing methodologies: Logistic Regression (LR) and the Training of Truth and Polarity Direction (TTPD). These tables serve as a supplement to the graphical representations shown in Figures 2 and 3 in the main body of the paper, offering precise numerical values for the average accuracy at each layer, potentially including standard deviations as indicated in the original tables. This detailed data allows for a granular understanding of how linearly decodable the model's instructed "True"/"False" output is at various depths within the network. By providing these comprehensive figures, researchers can more meticulously examine how different model architectures, instruction types, and probing techniques influence the predictability of representations across layers, and further verify the conclusions drawn in the main text regarding key layers where significant representational shifts or peak predictability occurs.

\begin{table*}[ht]
\centering
\begin{adjustbox}{max width=\textwidth}
\begin{tabular}{c|cc|cc|cc}
\toprule
\textbf{Layer} & \multicolumn{2}{c|}{\textbf{Truthful}} & \multicolumn{2}{c|}{\textbf{Neutral}} & \multicolumn{2}{c}{\textbf{Deceptive}} \\
& \textbf{TTPD} & \textbf{LR} & \textbf{TTPD} & \textbf{LR} & \textbf{TTPD} & \textbf{LR} \\
\midrule
1 & 50.69 ± 1.10 & 49.50 ± 0.00 & 50.43 ± 0.65 & 49.50 ± 0.00 & 50.63 ± 1.00 & 49.50 ± 0.00 \\
2 & 50.17 ± 0.35 & 49.56 ± 0.35 & 50.36 ± 0.66 & 49.48 ± 0.11 & 50.16 ± 0.36 & 49.48 ± 0.17 \\
3 & 50.07 ± 0.08 & 49.64 ± 0.27 & 50.21 ± 0.19 & 49.79 ± 0.43 & 50.20 ± 0.36 & 49.76 ± 0.45 \\
4 & 50.23 ± 0.31 & 50.62 ± 0.88 & 50.49 ± 0.50 & 50.74 ± 0.96 & 50.14 ± 0.24 & 50.11 ± 0.68 \\
5 & 49.83 ± 0.63 & 51.20 ± 0.64 & 49.88 ± 0.40 & 51.08 ± 0.83 & 49.91 ± 0.56 & 50.95 ± 0.62 \\
6 & 49.91 ± 0.55 & 50.92 ± 0.18 & 49.88 ± 0.47 & 50.93 ± 0.28 & 49.87 ± 0.24 & 50.84 ± 0.33 \\
7 & 52.38 ± 1.31 & 52.02 ± 0.64 & 51.58 ± 1.48 & 52.63 ± 1.06 & 51.01 ± 1.70 & 52.52 ± 1.12 \\
8 & 52.90 ± 1.26 & 52.13 ± 0.61 & 52.04 ± 0.86 & 52.94 ± 1.02 & 52.46 ± 0.75 & 52.12 ± 0.67 \\
9 & 58.66 ± 1.96 & 53.09 ± 1.60 & 57.93 ± 1.99 & 56.53 ± 3.05 & 57.77 ± 2.41 & 53.88 ± 2.34 \\
10 & 59.58 ± 4.57 & 61.37 ± 3.67 & 57.23 ± 4.96 & 62.20 ± 4.01 & 58.79 ± 3.42 & 62.05 ± 3.08 \\
11 & 68.41 ± 0.74 & 57.16 ± 3.11 & 64.66 ± 2.49 & 59.57 ± 3.65 & 60.11 ± 3.96 & 57.07 ± 3.38 \\
12 & 69.69 ± 0.63 & 63.11 ± 3.09 & 69.75 ± 0.83 & 62.33 ± 4.54 & 69.13 ± 0.73 & 58.76 ± 3.67 \\
13 & 77.20 ± 0.23 & 65.37 ± 4.95 & 80.10 ± 0.24 & 62.92 ± 5.64 & 73.43 ± 0.83 & 62.25 ± 4.66 \\
14 & 78.22 ± 0.23 & 75.70 ± 2.88 & 82.25 ± 0.53 & 71.66 ± 4.21 & 79.29 ± 0.30 & 70.44 ± 3.74 \\
15 & 80.04 ± 0.47 & 75.97 ± 3.46 & 84.86 ± 0.47 & 75.49 ± 3.53 & 80.42 ± 0.13 & 70.07 ± 6.07 \\
16 & 82.58 ± 0.60 & 77.15 ± 4.43 & 85.62 ± 0.14 & 74.91 ± 4.07 & 79.23 ± 0.11 & 71.33 ± 4.35 \\
17 & 83.62 ± 0.41 & 76.97 ± 3.27 & 85.42 ± 0.15 & 74.03 ± 5.79 & 79.80 ± 0.29 & 69.46 ± 6.76 \\
18 & 83.55 ± 0.35 & 74.53 ± 4.70 & 85.01 ± 0.30 & 74.43 ± 4.92 & 76.25 ± 0.35 & 67.11 ± 6.53 \\
19 & 83.24 ± 0.22 & 73.62 ± 5.45 & 83.75 ± 0.34 & 73.71 ± 4.70 & 78.69 ± 0.18 & 67.65 ± 5.26 \\
20 & 83.07 ± 0.41 & 74.90 ± 4.18 & 83.62 ± 0.22 & 71.85 ± 4.34 & 78.69 ± 0.48 & 66.16 ± 7.26 \\
21 & 82.84 ± 0.27 & 69.61 ± 5.96 & 83.41 ± 0.36 & 76.20 ± 4.87 & 79.50 ± 0.34 & 66.87 ± 6.48 \\
22 & 82.53 ± 0.35 & 71.29 ± 5.17 & 83.61 ± 0.21 & 73.31 ± 5.80 & 79.22 ± 0.23 & 66.44 ± 7.18 \\
23 & 82.25 ± 0.29 & 73.93 ± 5.95 & 83.39 ± 0.24 & 72.20 ± 4.89 & 79.06 ± 0.37 & 67.68 ± 6.19 \\
24 & 82.11 ± 0.37 & 71.52 ± 5.36 & 83.22 ± 0.24 & 72.48 ± 6.43 & 78.54 ± 0.33 & 68.94 ± 5.61 \\
25 & 82.10 ± 0.29 & 73.07 ± 6.04 & 83.30 ± 0.33 & 72.60 ± 6.11 & 78.24 ± 0.19 & 67.35 ± 6.50 \\
26 & 81.96 ± 0.29 & 71.62 ± 6.31 & 83.26 ± 0.19 & 73.14 ± 5.08 & 78.33 ± 0.26 & 70.58 ± 4.26 \\
27 & 81.55 ± 0.28 & 71.34 ± 6.16 & 82.97 ± 0.23 & 73.90 ± 5.83 & 77.45 ± 0.56 & 69.06 ± 5.07 \\
28 & 81.42 ± 0.33 & 73.38 ± 5.56 & 82.90 ± 0.24 & 75.66 ± 5.13 & 76.81 ± 0.30 & 68.20 ± 5.64 \\
29 & 81.36 ± 0.18 & 75.02 ± 3.82 & 82.70 ± 0.25 & 73.47 ± 4.71 & 76.43 ± 0.38 & 68.37 ± 6.25 \\
30 & 81.31 ± 0.22 & 70.84 ± 6.26 & 82.87 ± 0.34 & 72.73 ± 4.14 & 76.51 ± 0.60 & 69.63 ± 7.97 \\
31 & 81.62 ± 0.23 & 71.57 ± 5.66 & 82.89 ± 0.33 & 71.96 ± 5.07 & 76.17 ± 0.71 & 63.63 ± 7.32 \\
32 & 81.57 ± 0.33 & 64.56 ± 4.66 & 83.87 ± 0.19 & 72.36 ± 5.25 & 77.86 ± 1.23 & 68.55 ± 5.39 \\
% Add more rows as needed
\hline
\end{tabular}
\end{adjustbox}
\caption{Layer-wise probing accuracy for Llama3.1-8B-IT across truthful, neutral, and deceptive prompts using TTPD and LR.}
\label{tab:layer-wise_prob_llama}
\end{table*}

\begin{table*}[ht]
\centering
\begin{adjustbox}{max width=\textwidth}
\begin{tabular}{c|cc|cc|cc}
\toprule
\textbf{Layer} & \multicolumn{2}{c|}{\textbf{Truthful}} & \multicolumn{2}{c|}{\textbf{Neutral}} & \multicolumn{2}{c}{\textbf{Deceptive}} \\
& \textbf{TTPD} & \textbf{LR} & \textbf{TTPD} & \textbf{LR} & \textbf{TTPD} & \textbf{LR} \\
\midrule
1 & 50.47 ± 0.59 & 50.69 ± 0.51 & 51.03 ± 0.91 & 50.75 ± 0.61 & 50.54 ± 0.74 & 50.18 ± 0.47 \\
2 & 51.13 ± 0.73 & 50.55 ± 0.68 & 51.45 ± 1.29 & 50.15 ± 0.66 & 50.73 ± 0.76 & 50.41 ± 0.80 \\
3 & 50.80 ± 1.16 & 50.51 ± 0.70 & 51.68 ± 0.99 & 50.45 ± 0.65 & 51.07 ± 0.76 & 50.97 ± 0.77 \\
4 & 51.24 ± 0.55 & 50.55 ± 0.69 & 51.22 ± 1.00 & 50.49 ± 0.49 & 50.98 ± 0.67 & 49.64 ± 0.61 \\
5 & 51.04 ± 0.56 & 50.95 ± 0.51 & 50.84 ± 0.72 & 51.14 ± 0.30 & 51.40 ± 0.71 & 51.38 ± 0.60 \\
6 & 51.50 ± 0.62 & 50.79 ± 0.39 & 50.75 ± 0.55 & 51.20 ± 0.58 & 51.35 ± 0.78 & 50.88 ± 0.35 \\
7 & 48.31 ± 1.22 & 49.89 ± 0.82 & 48.09 ± 0.75 & 49.69 ± 1.17 & 48.17 ± 1.35 & 49.42 ± 0.90 \\
8 & 49.17 ± 0.91 & 50.68 ± 1.09 & 49.75 ± 0.84 & 51.09 ± 0.63 & 48.43 ± 0.99 & 50.71 ± 0.67 \\
9 & 52.31 ± 0.72 & 51.17 ± 0.71 & 52.45 ± 1.02 & 51.84 ± 1.05 & 52.99 ± 0.50 & 51.13 ± 0.62 \\
10 & 52.74 ± 0.86 & 51.85 ± 0.58 & 51.93 ± 0.66 & 52.31 ± 1.02 & 53.31 ± 0.36 & 51.90 ± 0.54 \\
11 & 52.98 ± 1.08 & 52.99 ± 0.78 & 51.92 ± 0.33 & 53.58 ± 1.63 & 53.37 ± 0.89 & 51.96 ± 0.67 \\
12 & 51.72 ± 0.33 & 52.40 ± 0.87 & 51.74 ± 0.32 & 53.76 ± 0.87 & 53.00 ± 1.27 & 52.47 ± 0.65 \\
13 & 52.63 ± 0.94 & 53.56 ± 0.99 & 52.11 ± 0.43 & 53.38 ± 1.52 & 52.95 ± 1.22 & 52.31 ± 0.78 \\
14 & 53.74 ± 1.31 & 53.92 ± 1.09 & 54.00 ± 1.97 & 53.80 ± 1.49 & 53.28 ± 1.30 & 52.44 ± 0.52 \\
15 & 54.80 ± 1.16 & 57.56 ± 2.03 & 54.32 ± 1.83 & 54.40 ± 1.81 & 54.11 ± 0.81 & 55.54 ± 1.31 \\
16 & 57.54 ± 1.84 & 62.03 ± 2.06 & 56.46 ± 2.46 & 57.29 ± 3.36 & 57.61 ± 0.76 & 59.86 ± 1.87 \\
17 & 60.84 ± 1.98 & 65.56 ± 1.83 & 58.78 ± 2.10 & 64.45 ± 2.30 & 60.17 ± 1.41 & 62.68 ± 1.38 \\
18 & 66.32 ± 1.09 & 68.80 ± 1.28 & 64.44 ± 2.09 & 65.77 ± 2.68 & 64.41 ± 1.36 & 63.64 ± 2.78 \\
19 & 75.17 ± 0.34 & 70.32 ± 5.99 & 77.33 ± 0.96 & 65.93 ± 6.63 & 69.10 ± 0.86 & 69.75 ± 4.53 \\
20 & 78.82 ± 0.47 & 72.61 ± 6.49 & 82.78 ± 0.95 & 70.72 ± 4.39 & 76.47 ± 0.44 & 72.12 ± 6.80 \\
21 & 82.52 ± 0.20 & 72.38 ± 6.30 & 84.72 ± 0.28 & 69.97 ± 6.46 & 84.16 ± 0.51 & 76.32 ± 4.30 \\
22 & 83.85 ± 0.34 & 64.00 ± 6.21 & 76.57 ± 0.95 & 65.25 ± 6.95 & 83.50 ± 0.26 & 71.22 ± 5.84 \\
23 & 83.63 ± 0.36 & 71.50 ± 7.20 & 77.85 ± 0.98 & 71.69 ± 5.75 & 82.76 ± 0.33 & 69.36 ± 7.19 \\
24 & 85.27 ± 0.11 & 71.62 ± 7.54 & 83.21 ± 0.47 & 73.37 ± 6.95 & 87.13 ± 0.75 & 70.12 ± 7.54 \\
25 & 85.40 ± 0.23 & 73.09 ± 6.16 & 84.82 ± 0.28 & 72.89 ± 6.10 & 84.49 ± 0.74 & 72.56 ± 7.26 \\
26 & 86.06 ± 0.20 & 75.38 ± 6.02 & 86.14 ± 0.44 & 68.37 ± 6.16 & 85.15 ± 0.65 & 71.84 ± 7.07 \\
27 & 85.77 ± 0.39 & 79.72 ± 4.74 & 84.89 ± 0.39 & 71.21 ± 5.94 & 86.70 ± 0.18 & 74.29 ± 5.96 \\
28 & 85.82 ± 0.17 & 79.99 ± 5.13 & 85.01 ± 0.28 & 76.32 ± 6.01 & 86.39 ± 0.49 & 73.53 ± 7.26 \\
29 & 85.90 ± 0.13 & 80.06 ± 5.60 & 85.26 ± 0.27 & 78.20 ± 5.66 & 83.64 ± 0.45 & 72.23 ± 7.11 \\
30 & 85.58 ± 0.21 & 79.15 ± 5.15 & 84.56 ± 0.28 & 74.26 ± 5.50 & 79.60 ± 0.51 & 74.41 ± 5.91 \\
31 & 85.37 ± 0.19 & 76.35 ± 6.26 & 84.64 ± 0.26 & 75.32 ± 6.12 & 75.75 ± 0.80 & 75.85 ± 7.14 \\
32 & 85.68 ± 0.22 & 77.95 ± 6.29 & 84.98 ± 0.23 & 77.45 ± 4.96 & 74.61 ± 0.89 & 71.97 ± 7.13 \\
% Add more rows as needed
\hline
\end{tabular}
\end{adjustbox}
\caption{Layer-wise probing accuracy for Gemma2-9B-IT across truthful, neutral, and deceptive prompts using TTPD and LR.}
\label{tab:layer-wise_prob_gemma}
\end{table*}

\FloatBarrier

\section{PCA Visualization Results}
\label{appendix:pca_plots}
This section presents supplementary PCA visualizations to further illustrate the global geometry of the models' internal activations under different instructional prompts. As discussed in the main text, PCA is employed to project the high-dimensional hidden state activations (\(x_l\)) onto a 2D space, primarily for illustrative purposes. These visualizations help in assessing the separability of internal states corresponding to "True" and "False" outputs across Neutral, Truthful, and Deceptive conditions.

\begin{figure}[t]
    \centering
    \includegraphics[width=\linewidth]{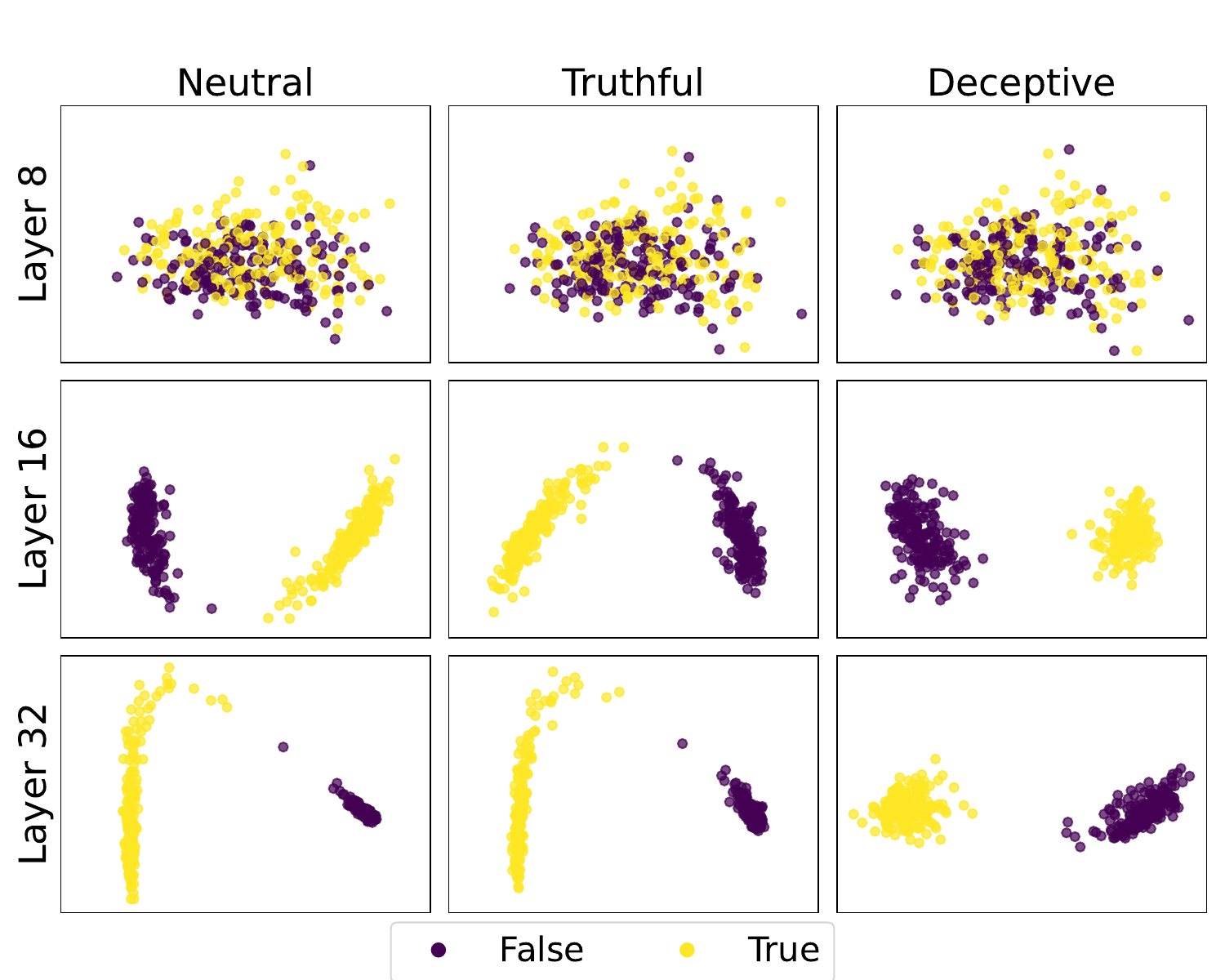}
    \caption{Layer-wise PCA visualization for LLaMA-3.1-8B-Instruct across Neutral, Truthful, and Deceptive Prompts on \textbf{\texttt{sp\_en\_trans}}}
    \label{fig:pca_llama3_8b_sp_en_trans}
\end{figure}

\begin{figure}[t]
    \centering
    \includegraphics[width=\linewidth]{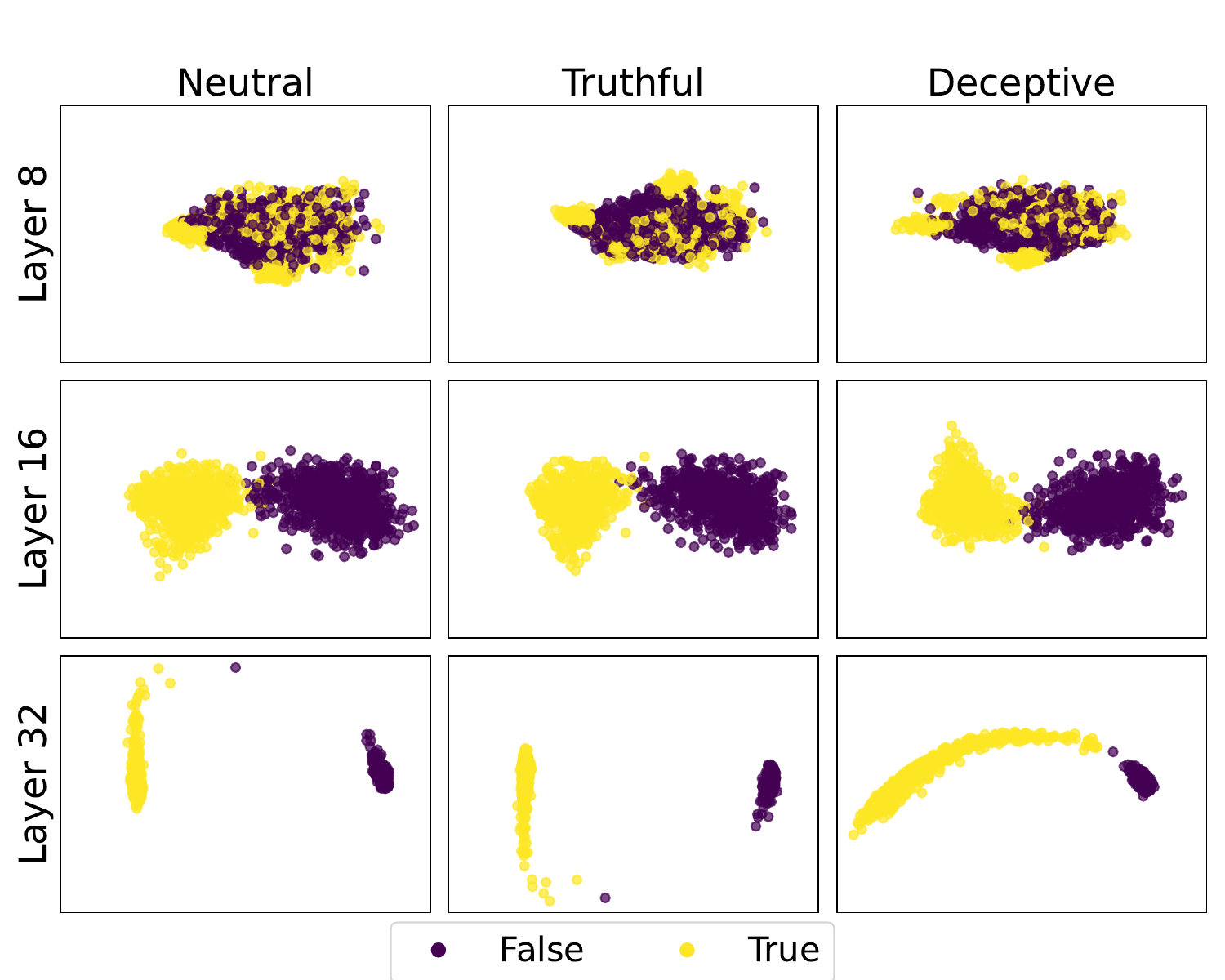}
    \caption{Layer-wise PCA visualization for Gemma-2-9B-Instruct on \textbf{\texttt{cities}}.}
    \label{fig:pca_gemma2_9b_cities}
\end{figure}

\begin{figure}[t]
    \centering
    \includegraphics[width=\linewidth]{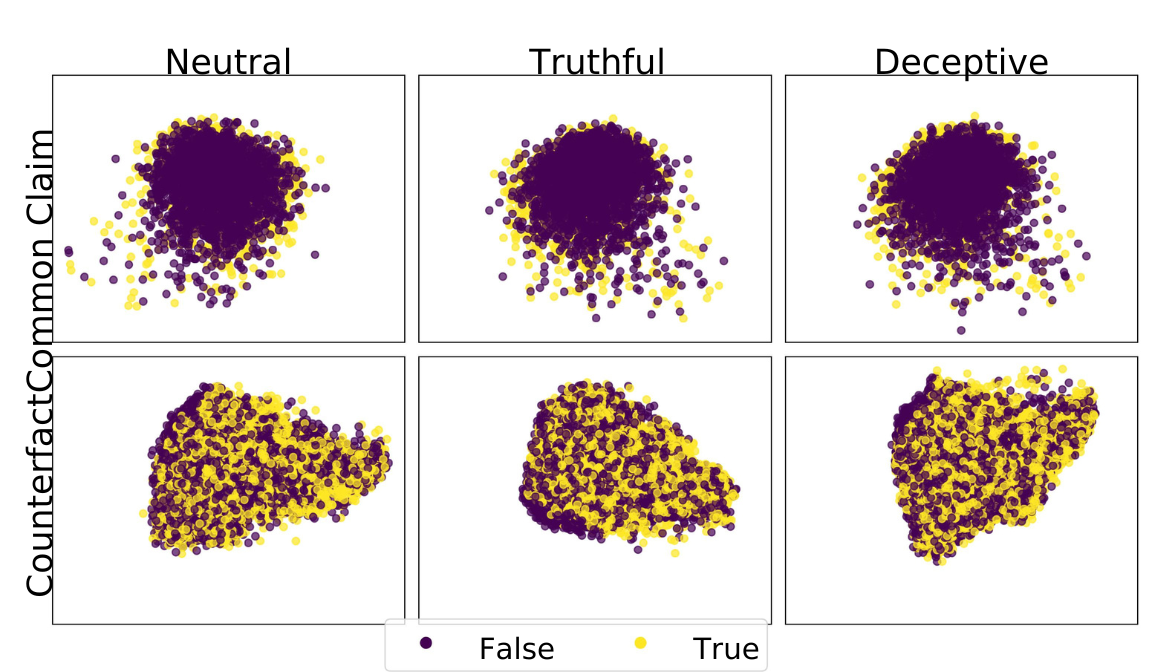}
    \caption{Layer-wise PCA visualization (Component 1 vs. Component 2) for Gemma-2-9B-Instruct on two complex datasets: \textbf{\texttt{common\_claim\_true\_false}} (top row) and \textbf{\texttt{counterfact\_true\_false}} (bottom row). Columns represent different instructional conditions: Neutral, Truthful, and Deceptive prompts. Visualizations are performed on \textbf{Layer 16}, the key layer identified for Gemma-2-9B-Instruct based on the probing accuracy peaks in Figure \ref{fig:logical_probing}. }
    \label{fig:pca_gemma2_9b_limit}
\end{figure}

The figures below provide additional examples beyond those in Section 3.2, showcasing these dynamics for both LLaMA-3.1-8B-Instruct and Gemma-2-9B-Instruct on various datasets. Specifically, Figure~\ref{fig:pca_llama3_8b_sp_en_trans} and Figure~\ref{fig:pca_gemma2_9b_cities} demonstrate the PCA results on curated datasets (e.g., \texttt{sp\_en\_trans} for LLaMA and \texttt{cities} for Gemma). These typically show a clearer separation between True/False clusters as observed in Figure 4 for the \texttt{cities} dataset with LLaMA. In contrast, Figure~\ref{fig:pca_llama_3.1_8b_limit} (which may correspond to Figure 5 in the main text showing \texttt{common\_claim} and \texttt{counterfact} for LLaMA) and Figure~\ref{fig:pca_gemma2_9b_limit} (showing similar complex datasets for Gemma, as in Figure 12) illustrate the challenges PCA faces with more complex, uncurated datasets where the True/False clusters often appear entangled due to feature superposition. These appendix figures offer a broader visual substantiation of how the geometric separability of truth-related representations can vary significantly with dataset complexity and model type.

\FloatBarrier

\section{SAE-based Layer-wise Feature Shift Analysis}
\label{appendix:sae_layer}
\begin{figure}[t]
    \centering
    \includegraphics[width=1\linewidth]{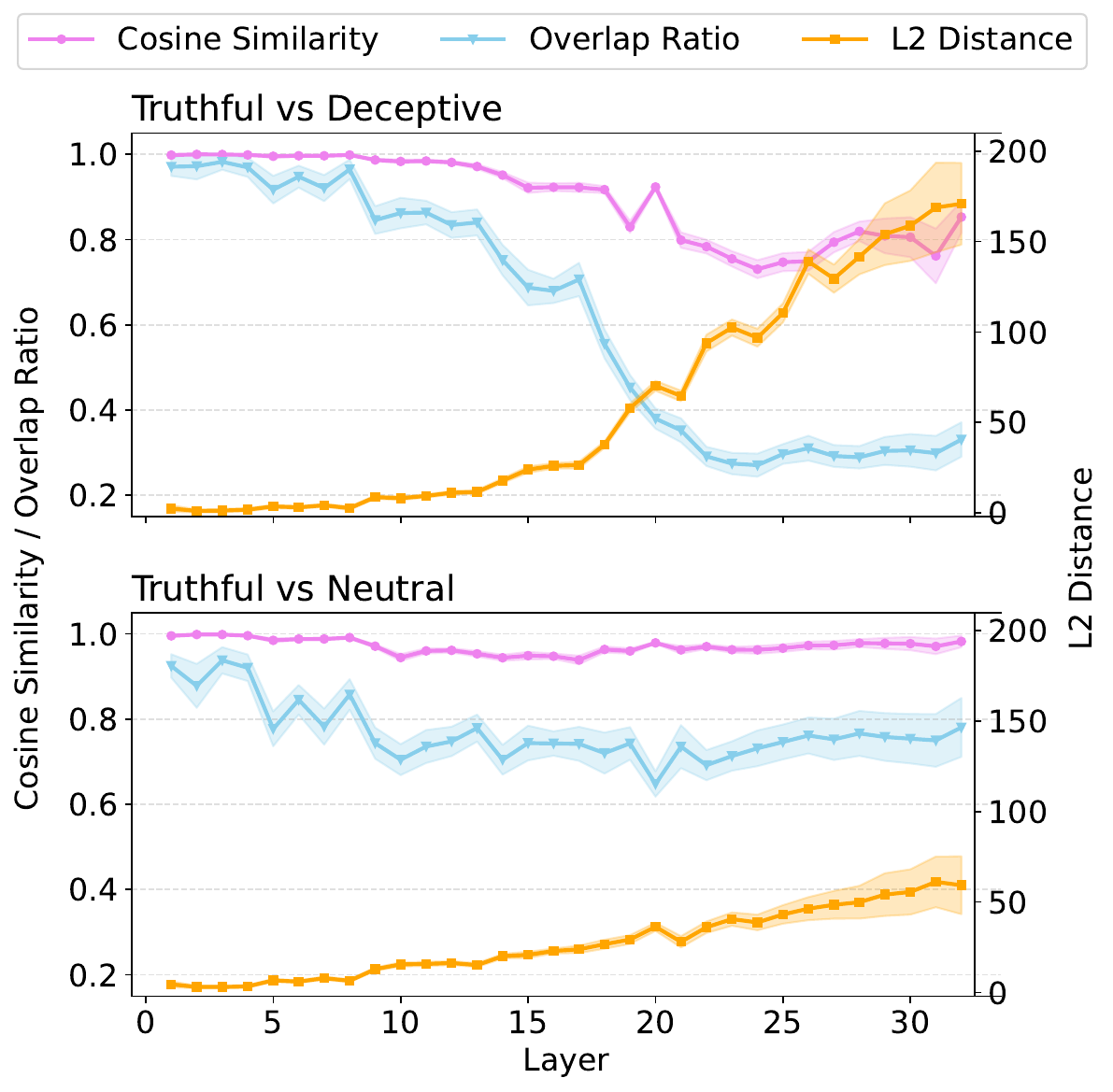}
    \caption{Layer-wise feature shift analysis for Gemma-2-9B-Instruct on \texttt{common\_claim\_true\_false}.}
    \label{fig:sae_gemma2_common}
\end{figure}

\begin{figure}[t]
    \centering
    \includegraphics[width=1\linewidth]{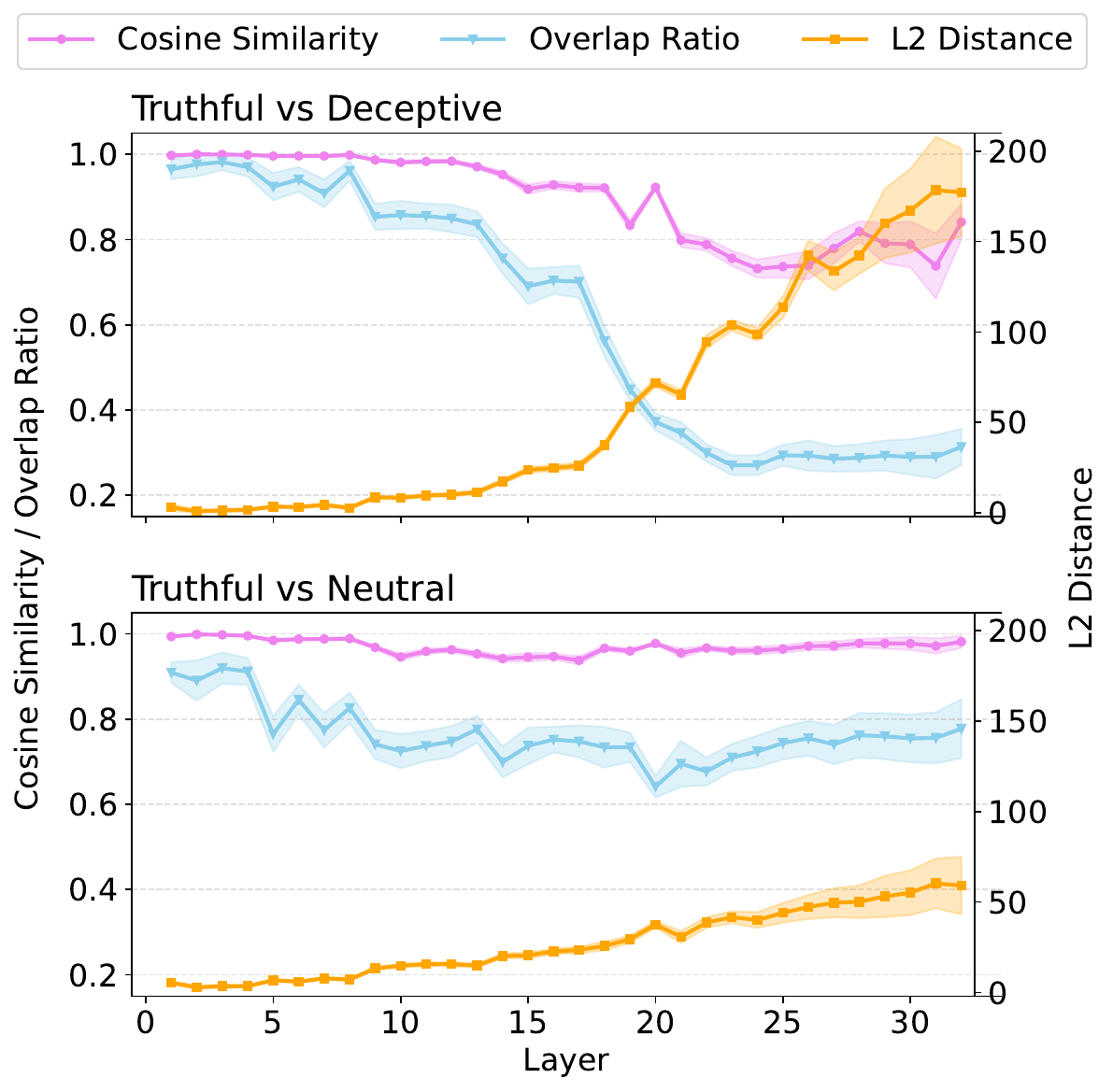}
    \caption{Layer-wise feature shift analysis for Gemma-2-9B-Instruct on \texttt{counterfact\_true\_false}.}
    \label{fig:sae_gemma2_counter}
\end{figure}

We analyze how sparse feature activations shift across layers under different instruction types using three metrics in SAE latent space: Cosine Similarity, Overlap Ratio, and L2 Distance. Figures~\ref{fig:sae_gemma2_common} and \ref{fig:sae_gemma2_counter} show results for the \texttt{common\_claim\_true\_false} and \texttt{counterfact\_true\_false} datasets using Gemma-2-9B-Instruct.

In both datasets, deceptive prompts induce strong mid-to-late layer shifts, especially between Layers 16 and 32. This is evidenced by the sharp rise in L2 distance and the corresponding drop in cosine similarity and overlap ratio when comparing truthful and deceptive inputs. The effect is most pronounced in \texttt{counterfact\_true\_false}, where overlap sharply declines post-Layer 16, indicating a reconfiguration of sparse feature sets. In contrast, shifts between truthful and neutral prompts remain small and gradual across all layers, suggesting that the major representational changes are deception-specific.

These results highlight a distinctive geometric transformation in the model’s latent representations under deceptive instructions and further motivate mid-layer analysis when identifying potential deception-sensitive features.
This pattern closely aligns with earlier findings from linear probing, where intermediate layers—especially around Layer 16—also showed peak decodability of the model’s intended "True"/"False" output across instruction types.


\FloatBarrier

\section{SAE-based Neuron-wise Feature Shift Analysis}
\label{appendix:sae_plots}
This appendix section provides additional visualizations to support the neuron-wise feature shift analysis detailed in main text. The core objective of this analysis is to move beyond global representational shifts and pinpoint specific SAE features that are most sensitive to the change from truthful to deceptive instructions. By examining individual SAE feature activations, we can gain a better understanding of how deception is encoded at the feature level.

\begin{figure*}[t]
  \includegraphics[width=0.96\linewidth]{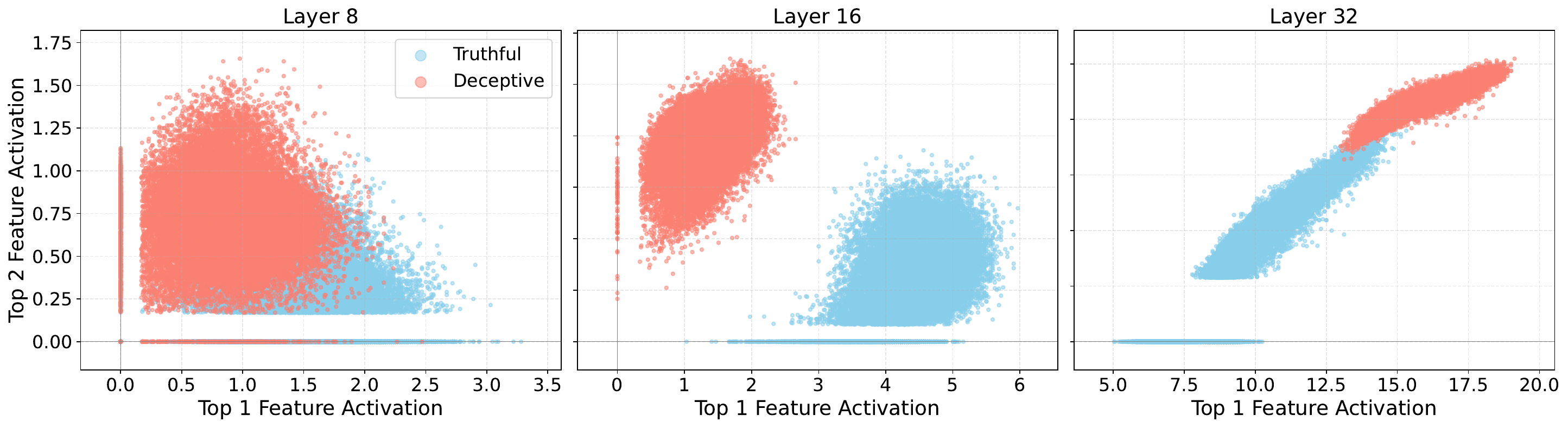}
  \caption{Neuron-by-neuron feature shift analysis for LLaMA-3.1-8B-Instruct on \texttt{counterfact\_true\_false}}
  \label{fig:neuron_scatter_counterfact}
\end{figure*}

\begin{figure*}[t]
  \centering
  % Layer 8 (a)
  \begin{minipage}{0.32\linewidth}
    \centering
    \includegraphics[width=\linewidth]{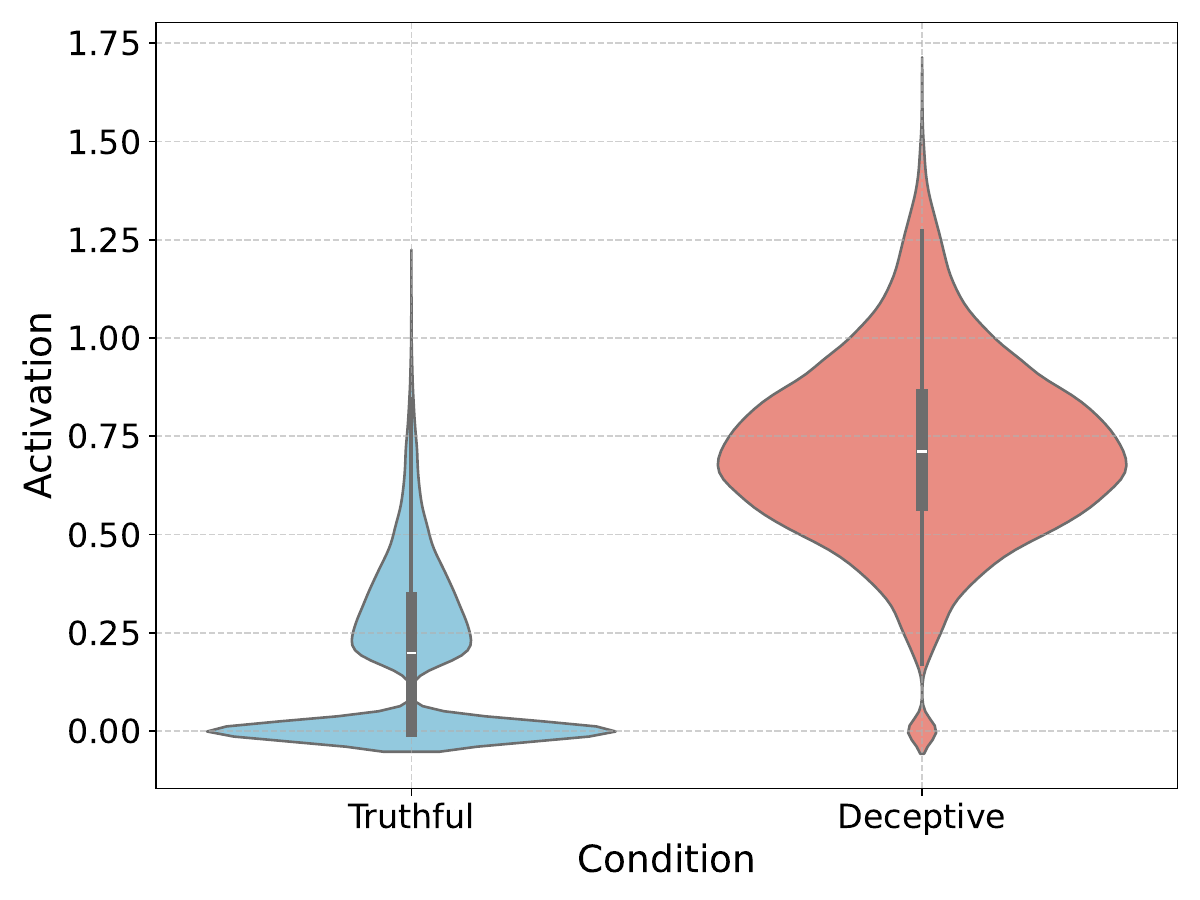}\\
    \includegraphics[width=\linewidth]{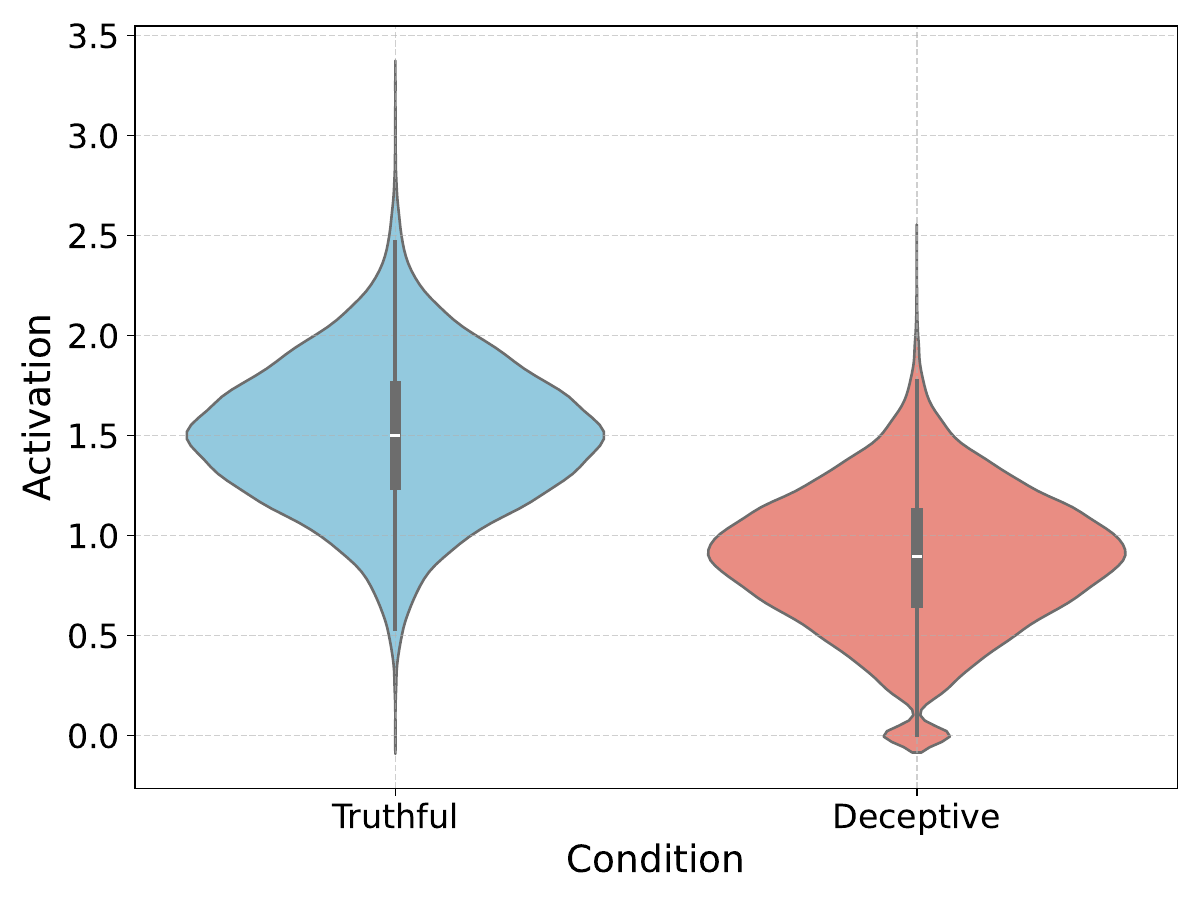}
    \subcaption{Layer 8}
  \end{minipage}\hfill
  % Layer 16 (b)
  \begin{minipage}{0.32\linewidth}
    \centering
    \includegraphics[width=\linewidth]{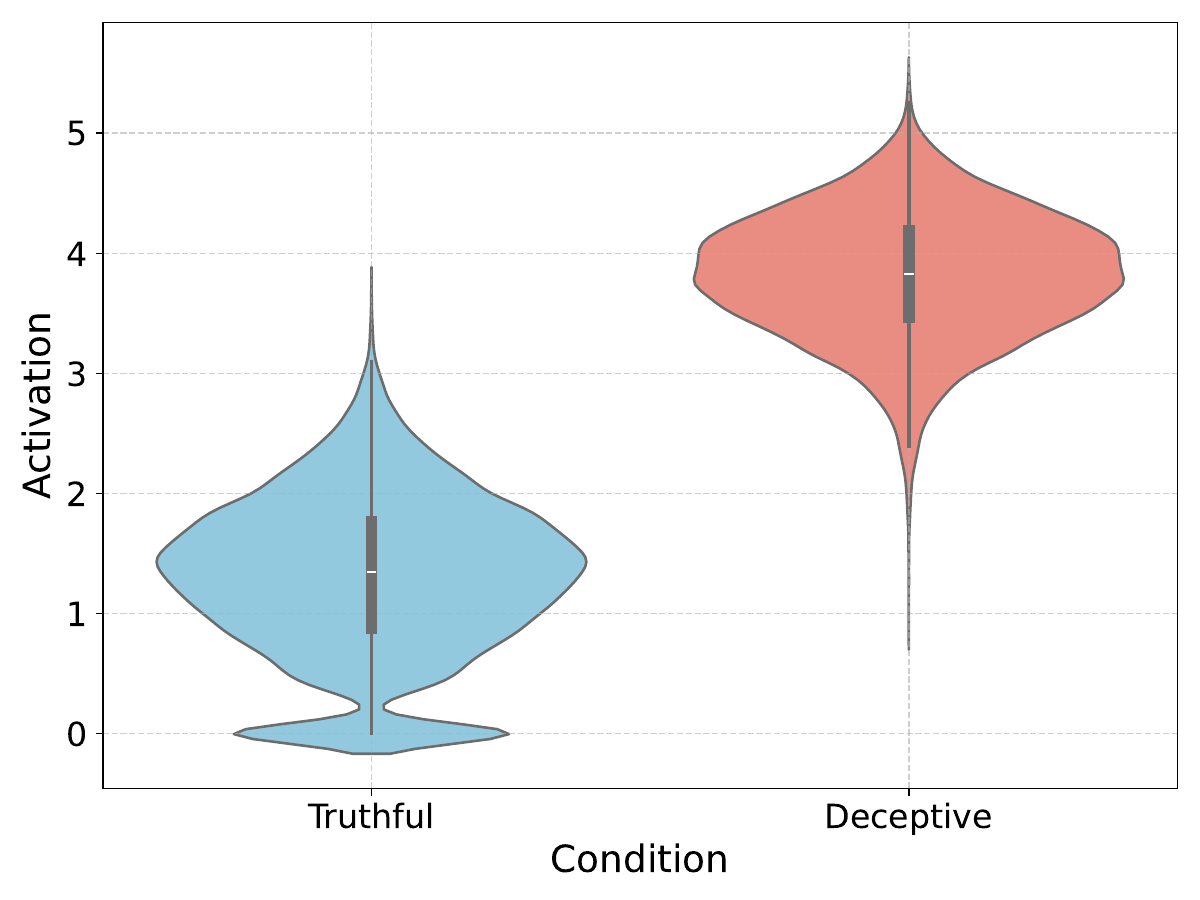}\\
    \includegraphics[width=\linewidth]{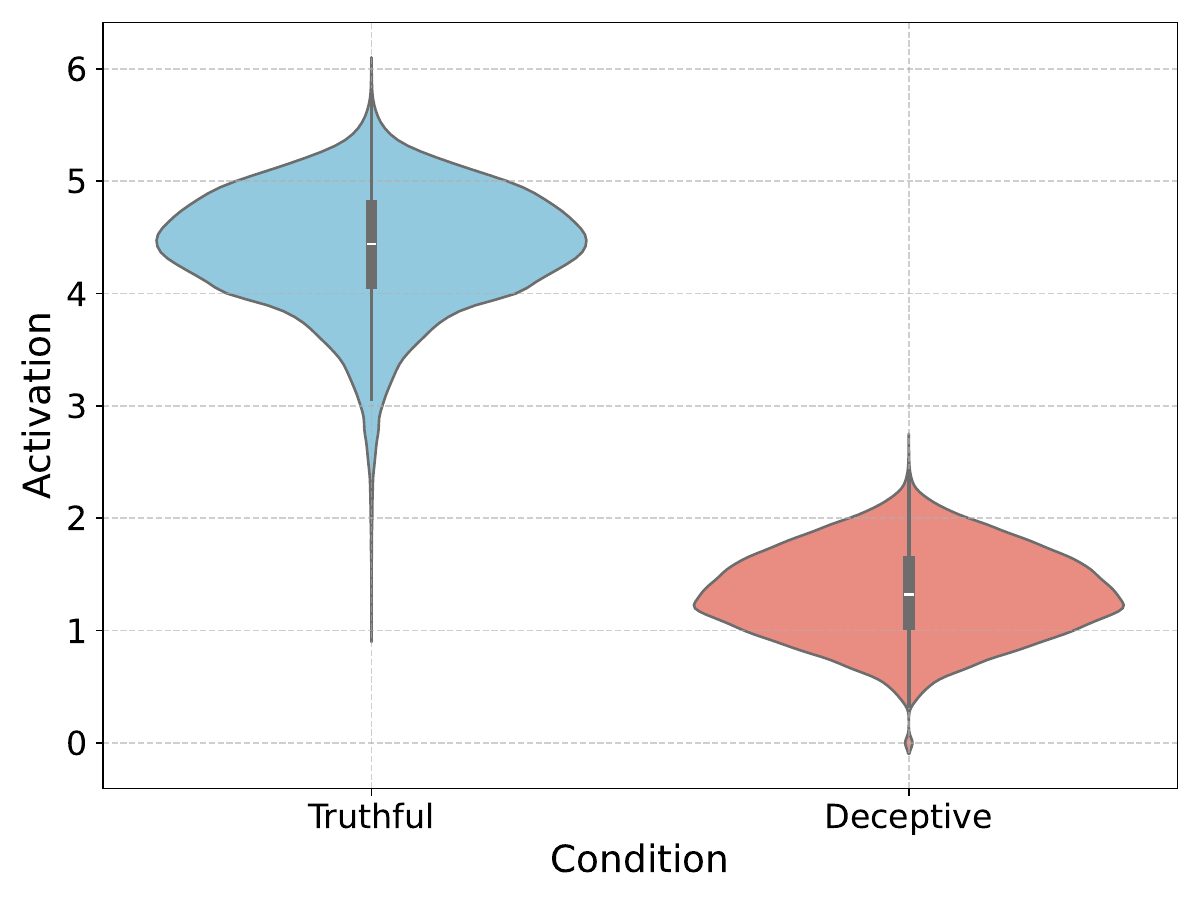}
    \subcaption{Layer 16}
  \end{minipage}\hfill
  % Layer 32 (c)
  \begin{minipage}{0.32\linewidth}
    \centering
    \includegraphics[width=\linewidth]{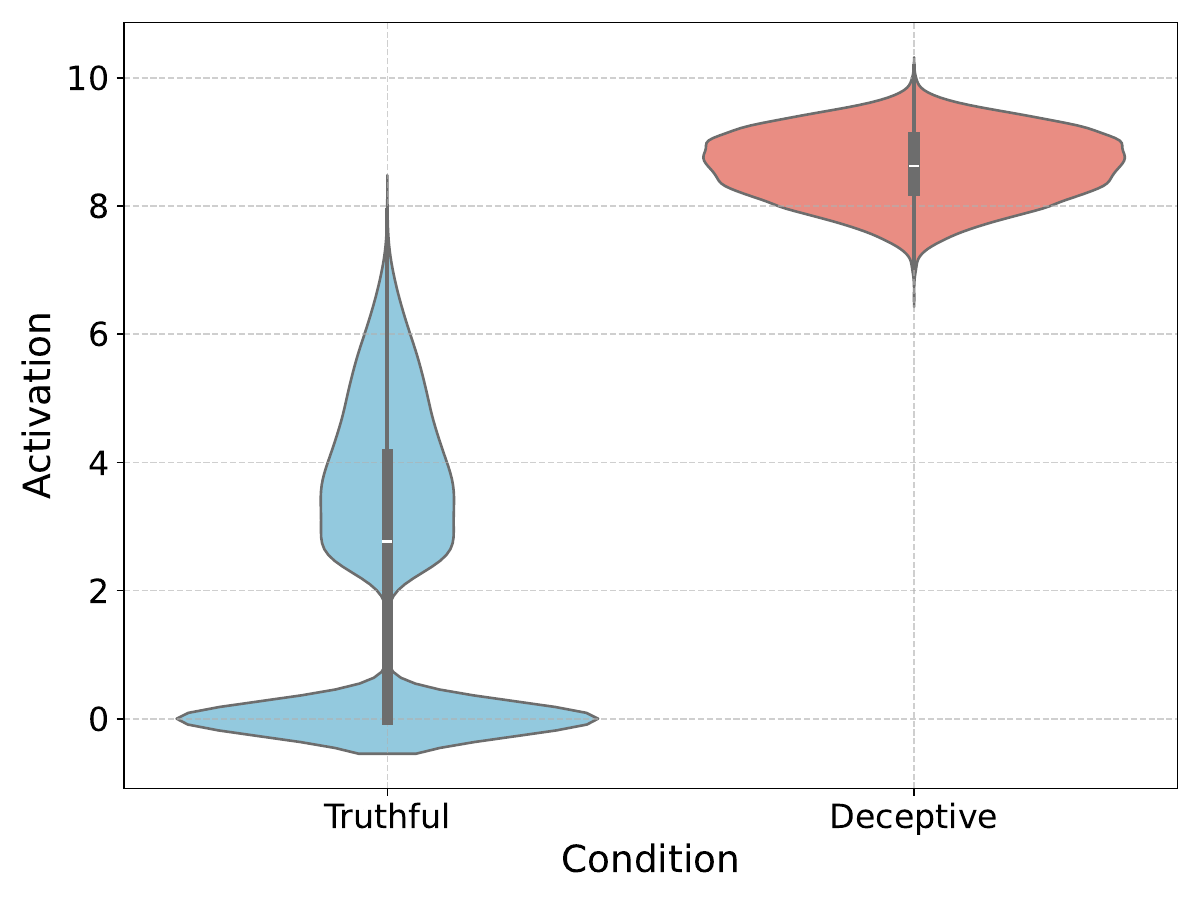}\\
    \includegraphics[width=\linewidth]{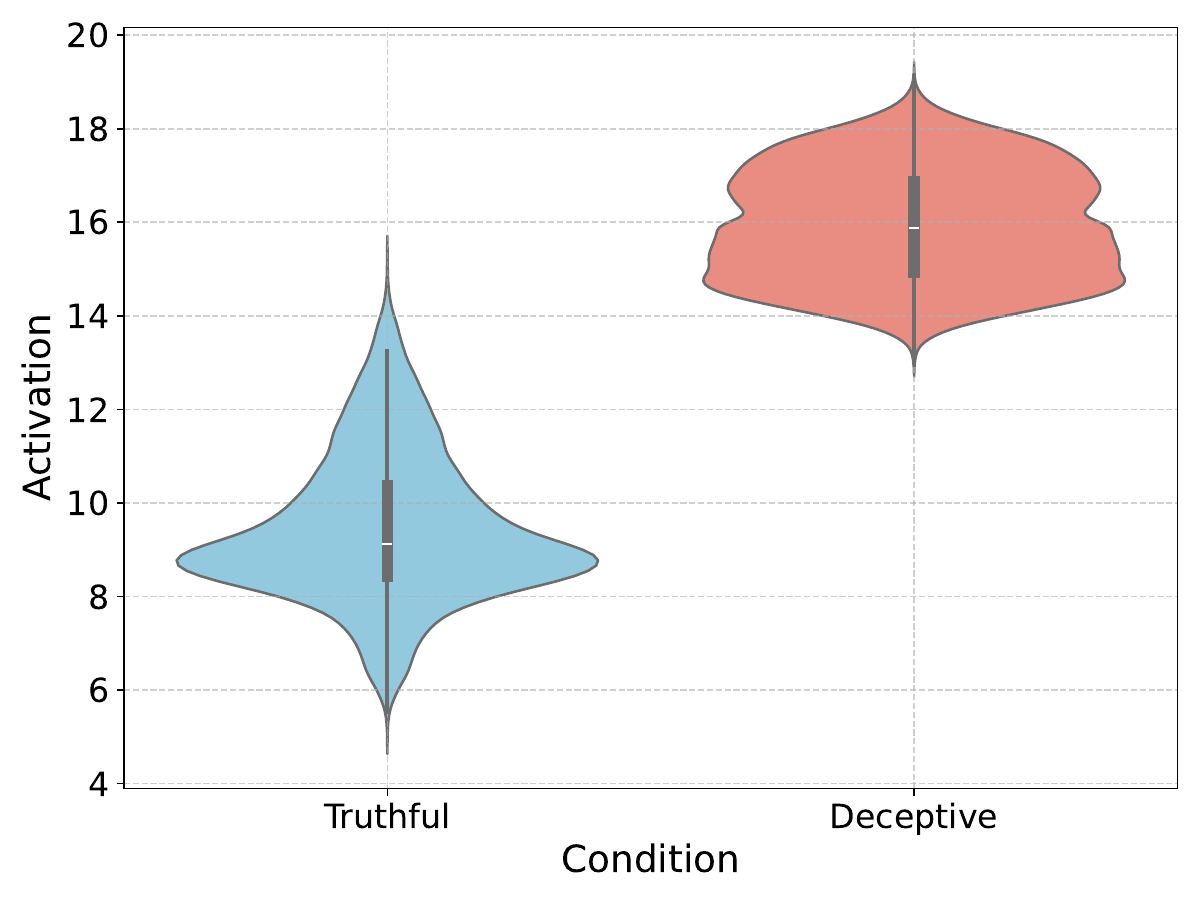}
    \subcaption{Layer 32}
  \end{minipage}

  \caption {
   Violin graph of LLaMA-3.1-8B-Instruct activations on the \texttt{counterfact\_true\_false}. The top row displays the activation distributions for the SAE feature most responsive to deceptive instructions (Top 1 feature), while the bottom row shows the distributions for the second most responsive feature (Top 2 feature), across layers 8 (a), 16 (b), and 32 (c).
}
  \label{fig:violin_counterfact}
\end{figure*}

The methodology involves identifying, for each layer, the sparse SAE features exhibiting the largest change in average activation when comparing the Deceptive condition to the Truthful condition. The figures presented in this appendix, such as scatter plots showing the activation of the top distinguishing features (similar to Figure~\ref{fig:sae_shift_commonclaim} for \texttt{common\_claim\_true\_false} but potentially for other datasets like \texttt{counterfact\_true\_false} as shown in Figure~\ref{fig:neuron_scatter_counterfact} and violin plots in Figure~\ref{fig:violin_counterfact} illustrating the distribution of these feature activations under truthful versus deceptive prompts, offer further evidence.

These supplementary visualizations help to reinforce the finding that a small subset of sparse features often displays a near-binary activation pattern—being highly active for one instruction type (e.g., truthful) and suppressed for the other (e.g., deceptive), or vice-versa. This detailed view corroborates the idea that these specific features act as "deception-associated features," playing a critical role in modulating the model's internal representation in response to deceptive instructions, often aligning with the mid- and late-layer SAE shift peaks identified globally in Figure~\ref{fig:sae_shift_cities} and Figure~\ref{fig:sae_shift_commonclaim}. The plots here may cover additional layers or datasets, providing a more comprehensive picture of this phenomenon.

\end{document}